\documentclass[letterpaper, 10 pt]{ieeeconf}  % Comment this line out if you need a4paper

\IEEEoverridecommandlockouts                              % This command is only needed if
\usepackage{graphics} % for pdf, bitmapped graphics files
\usepackage{epsfig} % for postscript graphics files
\usepackage{times} % assumes new font selection scheme installed
\usepackage{amsmath} % assumes amsmath package installed
\usepackage{bm}
\usepackage{amssymb}  % assumes amsmath package installed
\usepackage[ruled,vlined,linesnumbered]{algorithm2e}
\usepackage{mathtools}
\usepackage{caption}
\usepackage{subcaption}
\usepackage{breqn}
\DeclarePairedDelimiter{\ceil}{\lceil}{\rceil}
\usepackage{color}
\usepackage{graphicx}
\usepackage{diagbox}
\usepackage[utf8]{inputenc}
\usepackage[dvipsnames]{xcolor}
\usepackage{multicol}

\usepackage[english]{babel}

\usepackage{tabularx,booktabs}
\newcolumntype{C}{>{\centering\arraybackslash}X} % centered version of "X" type
\newtheorem{definition}{\bf Definition}
\newtheorem{theorem}{\bf Theorem}
\newtheorem{lemma}{\bf Lemma}
\newtheorem{problem}{\bf Problem}

\let\oldnl\nl% Store \nl in \oldnl
\newcommand{\nonl}{\renewcommand{\nl}{\let\nl\oldnl}}% Remove line number for one line
\makeatother

\usepackage{hyperref} 

\begin{document}

% }
\title{\LARGE \bf
 Risk-Aware Submodular Optimization for \\ Stochastic Travelling Salesperson Problem
}

\author{Rishab Balasubramanian, Lifeng Zhou, Pratap Tokekar, and P.B. Sujit %
\thanks{Rishab Balasubramanian is Research Associate at IISER Bhopal, Bhopal -- 462066, India (email: {\tt\small rishab.edu@gmail.com}).}
\thanks{Lifeng Zhou was with the Department of Electrical and Computer Engineering, Virginia Tech, Blacksburg, VA, USA when part of the work was completed. He is currently with the GRASP Laboratory, University of Pennsylvania, Philadelphia, PA, USA (email: {\tt\small lfzhou@seas.upenn.edu}).}
\thanks{Pratap Tokekar is Assistant Professor in the  Department of Computer Science at the University of Maryland, College Park, MD 20742, USA. (email: {\tt\small tokekar@umd.edu}).}
\thanks{P.B. Sujit is Associate Professor in the Department of Electrical Engineering and Computer Science, IISER Bhopal, Bhopal -- 462066, India (email: {\tt\small sujit@iiserb.ac.in}).}
\thanks{This work is supported in part by the National Science Foundation under Grant No. 1943368.}
}

\graphicspath{ {./figs/} }

\maketitle
\thispagestyle{empty}
\pagestyle{empty}

%%%%%%%%%%%%%%%%%%%%%%%%%%%%%%%%%%%%%%%%%%%%%%%%%%%%%%%%%%%%%%%%%%%%%%%%%%%%%%%%
\begin{abstract}

\textcolor{black}{We introduce a risk-aware variant of the Traveling Salesperson Problem (TSP), where the robot tour cost and reward have to be optimized simultaneously, while being subjected to uncertainty in both. We study the case where the rewards and the costs exhibit diminishing marginal gains, i.e., are submodular. Since the costs and the rewards are stochastic, we seek to maximize a risk metric known as Conditional-Value-at-Risk (CVaR) of the submodular function. We propose a Risk-Aware Greedy Algorithm (\texttt{RAGA}) to find an approximate solution for this problem. The approximation algorithm runs in polynomial time and is within a constant factor of the optimal and an additive term that depends on the value of optimal solution. We use the submodular function's curvature to improve approximation results further and verify the algorithm's performance through simulations.}

% We introduce a risk-aware solution to the Traveling Salesperson Problem (TSP), where the robot tour cost and reward have to be optimized simultaneously. The robot obtains reward along the edges in the graph, \textcolor{blue}{and is penalized proportional to the length of the resulting tour}. We study the case where the rewards and the costs exhibit diminishing marginal gains, i.e., are submodular. Unlike prior work, we focus on the scenario where the costs and the rewards are uncertain and seek to maximize the Conditional-Value-at-Risk (CVaR) metric of the submodular function. We propose a risk-aware greedy algorithm (\texttt{RAGA}) to find a bounded-approximation algorithm. The approximation algorithm runs in polynomial time and is within a constant factor of the optimal and an additive term that depends on the optimal solution. We use the submodular function's curvature to improve approximation results further and verify the algorithm's performance through simulations.

\end{abstract}

%%%%%%%%%%%%%%%%%%%%%%%%%%%%%%%%%%%%%%%%%%%%%%%%%%%%%%%%%%%%%%%%%%%%%%%%%%%%%%%%
\section{Introduction}
\begin{figure*}[t]
\begin{multicols}{4}
    \includegraphics[width=\linewidth]{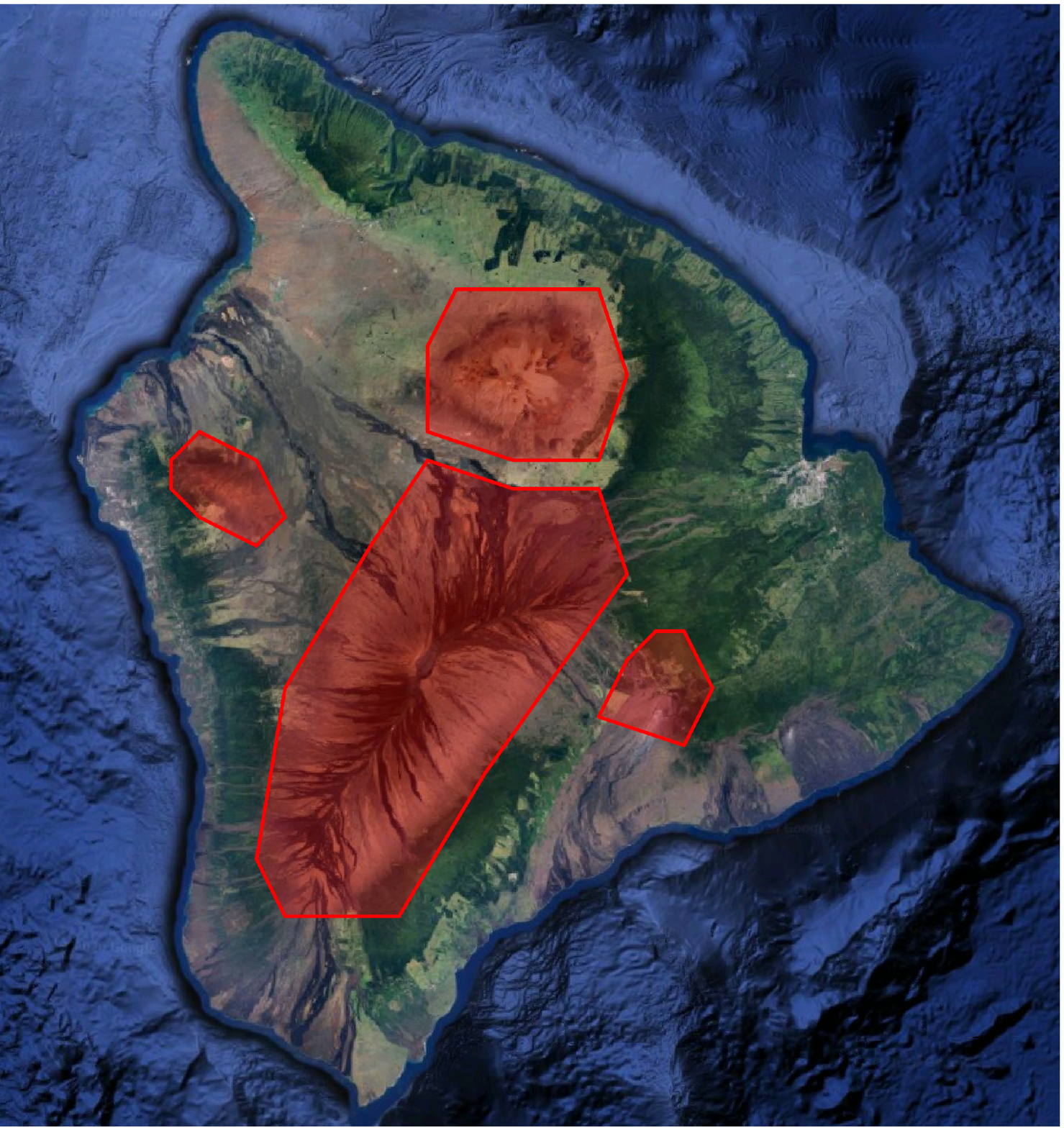}\subcaption{An image of an island, with the locations of active volcanoes highlighted in red}\label{fig:map_Hawaii}\par
    \includegraphics[width=\linewidth]{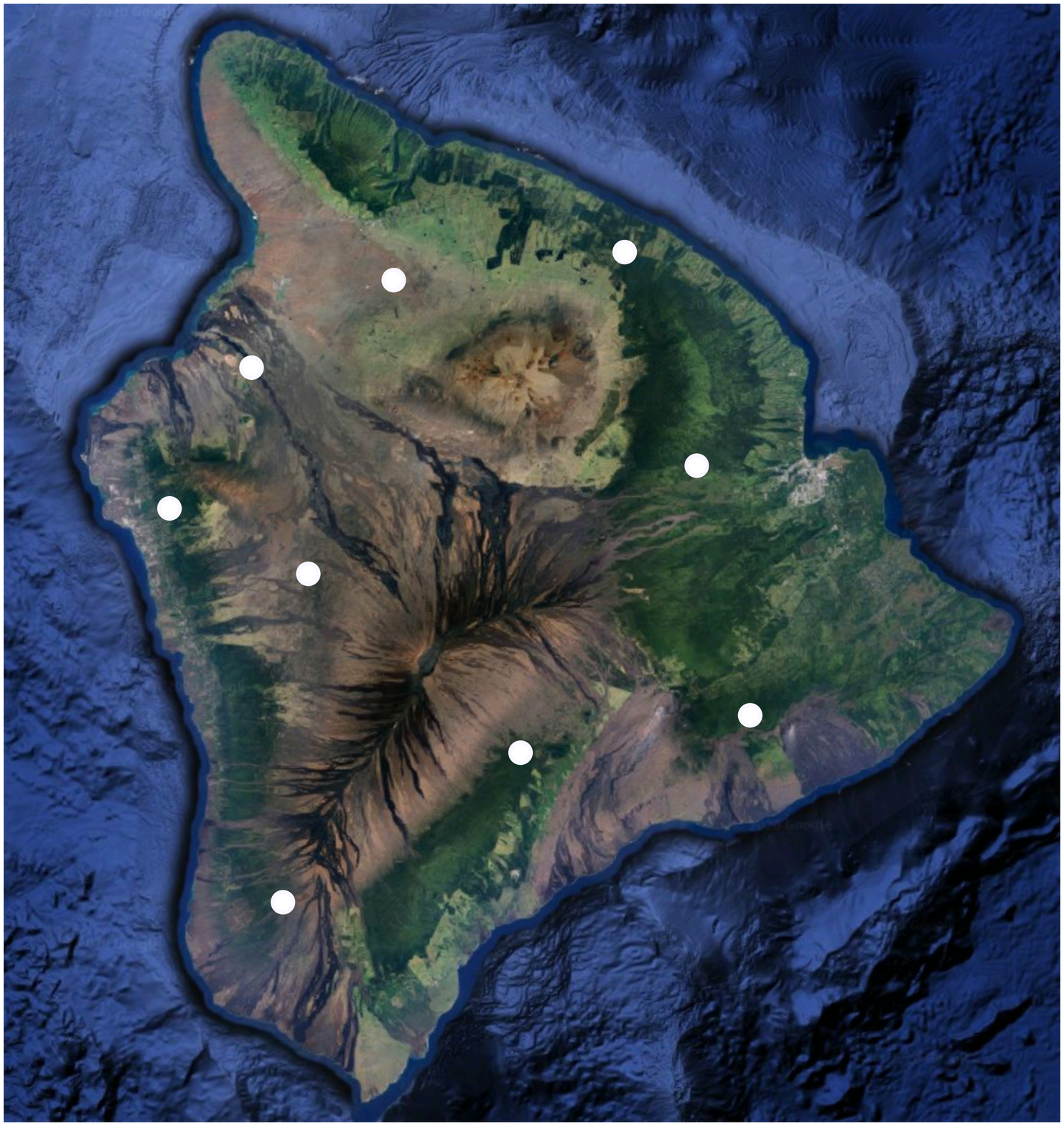}\subcaption{Key sites of surveillance surrounding the volcanic sites}\label{fig:key_points}\par
    \includegraphics[width=\linewidth]{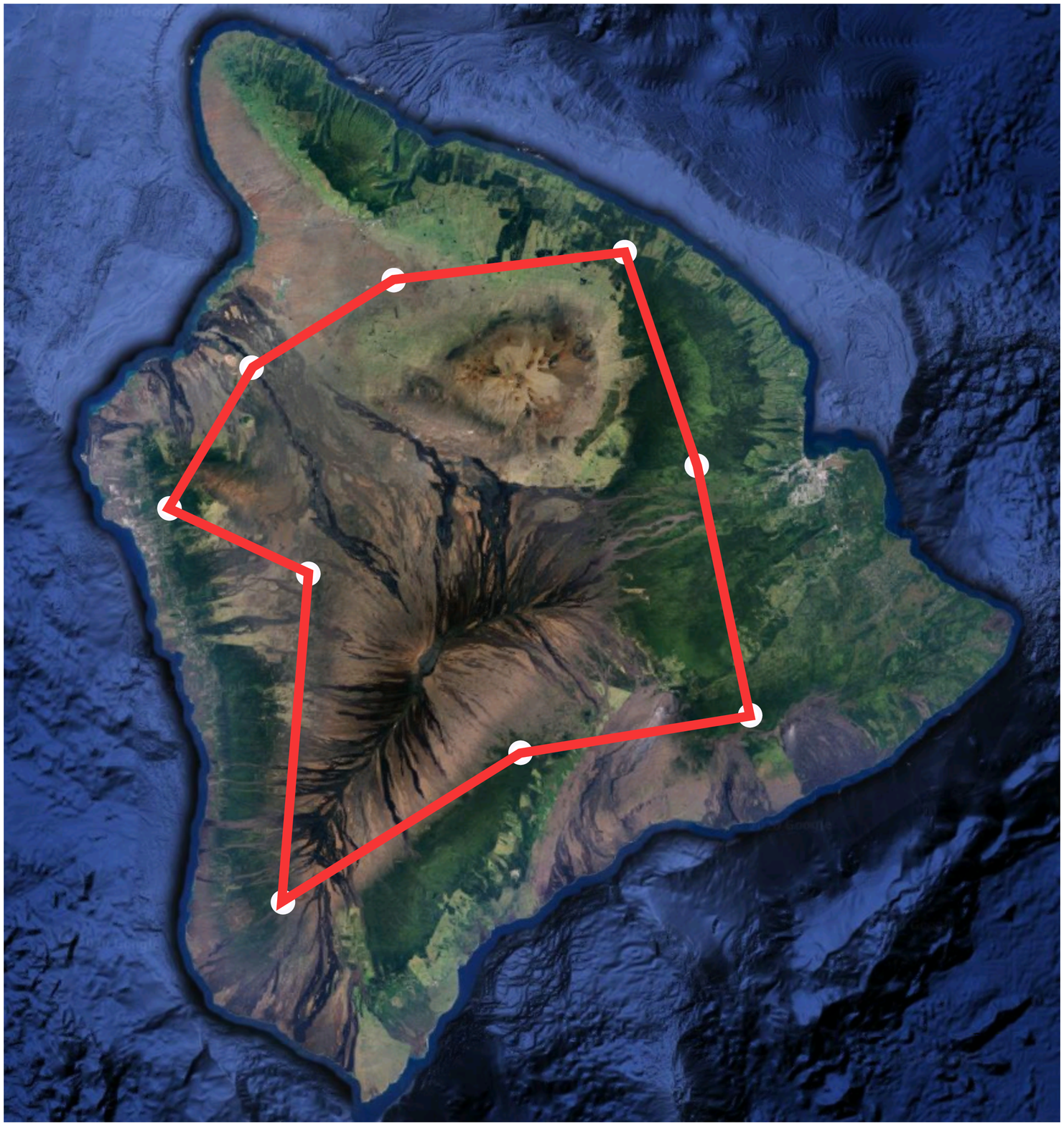}\subcaption{Tour with low cost and low reward}\label{fig:low_cost_low_reward}\par 
    \includegraphics[width=\linewidth]{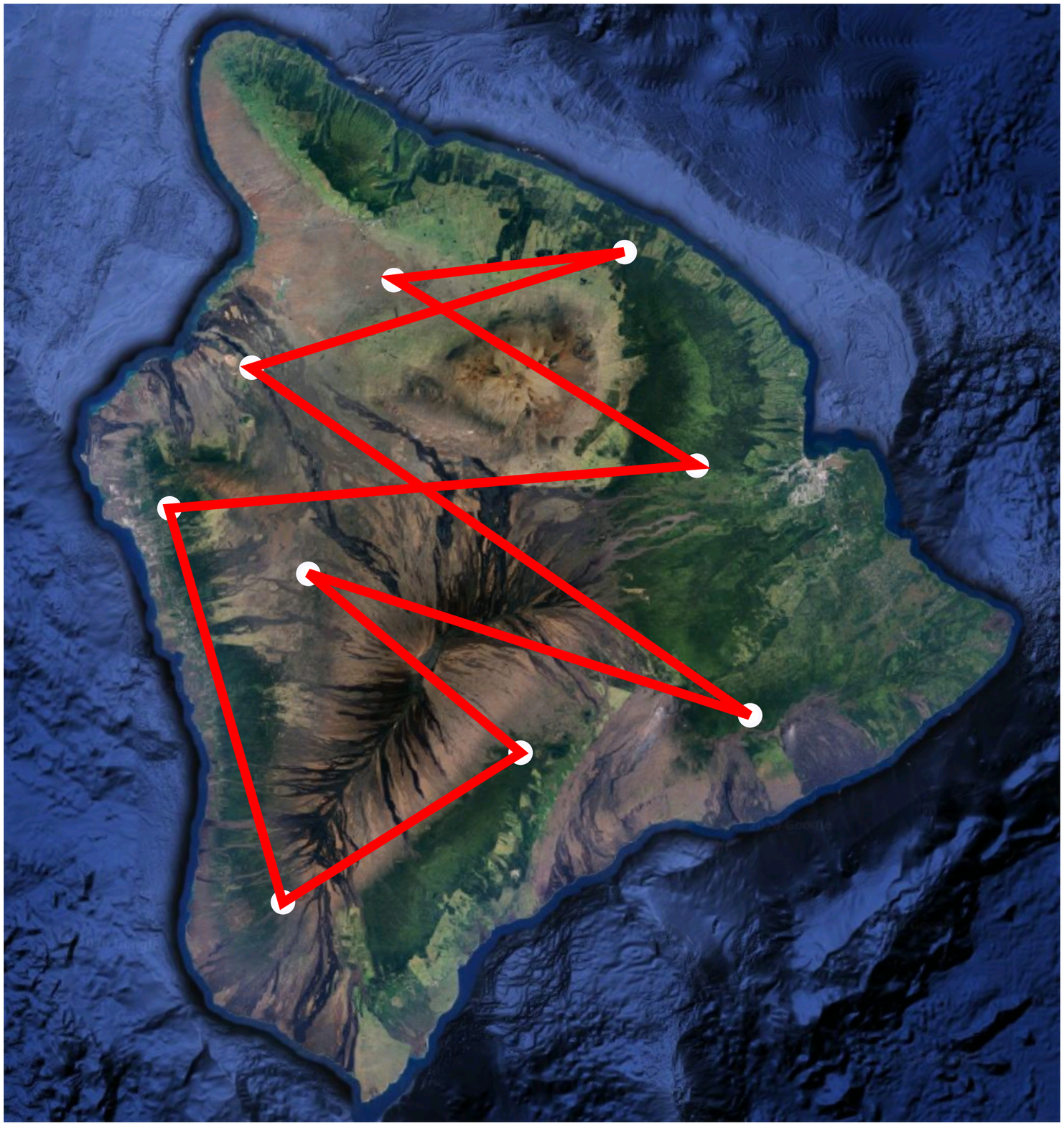}\subcaption{Tour with high cost and high reward}\label{fig:high_cost_high_reward}\par
\end{multicols}
\caption{An example of risk-aware tour selection for volcano monitoring in an island. A low-risk, low-reward monitoring tour avoids the more interesting region in the middle which the high-risk, high-reward tour covers.}\label{fig:Hawaii} 
%\lz{fig1 and 2 together}
\end{figure*}

%\pb {The role of CVaR is missing and the motivation is missing}
Determining an optimal tour to visit all the locations in a given set while minimizing/maximizing a metric is the classical Travelling Salesperson Problem (TSP) that finds applications in robotics, logistics, etc. However, there are several applications where the environment is dynamic and uncertain, \textcolor{black}{as a result of which classical approaches to solving the TSP are insufficient.} %that may affect the tour.
Examples include determining routes to visit active volcanic regions (where the activity has temporal variability) for obtaining scientific information (as shown in Fig. \ref{fig:Hawaii}); determining routes for a logistic delivery vehicle in dense urban regions with uncertain traffic, etc. In such scenarios, the risk due to uncertainty in the travel times and/or the rewards collected along the path needs to be considered while determining the tour. %In this paper, we address the risk--aware  multi-objective routing problem by developing a rewards--risk trade-off framework. 

Several approaches to stochastic TSP have been presented in the literature.  In \cite{perboli2017progressive}, a two-step process to convert the TSP to a multi-integer linear programming problem and then introduce a meta-heuristic based on the probabilistic hedging method proposed in \cite{rockafellar1991scenarios} is carried out. Paulin \cite{paulin2014convex} uses an extension of Stein's method for exchangeable pairs to approach the stochastic salesperson problem. In \cite{maity2015modified}, \cite{maity2016imprecise} and \cite{mukherjee2019constrained}, the authors develop a genetic algorithm to approach the uncertainty in TSP. In \cite{itani2007stochastic}, a constant factor approximation algorithm is developed for a Dubin's vehicle visiting all the points, with cost minimization being the main objective. In \cite{adler2016stochastic}, again an approximation algorithm is presented to minimize the time taken to visit the targets that are appearing stochastically in the environment. In all the above articles, the risk is not considered directly. 

We argue that a different approach is necessary in many risk-sensitive applications. Specifically, instead of optimizing the expected cost, optimizing a risk-sensitive measure may be more appropriate. In this paper, we focus on this case and present a risk-aware TSP formulation. To do so, we develop an approximation to the stochastic TSP with \textcolor{black}{the optimization objective} represented as a submodular function. The resulting algorithm takes as input a risk tolerance parameter, $\alpha$, and produces a tour that maximizes the expected behavior in the worst $\alpha$ percentile cases. Thus, the user can choose tours ranging from risk-neutral ($\alpha = 1$) to very conservative ($\alpha\approx 0$).

%\lz{it would be better to replace this paragraph by the work on TSP with uncertainty/stochastic tsp, then it can naturally transition to the next paragraph} \rb{addressed} \lz{exchange these 2 for a better organisation}
An important property of submodular functions are their diminishing marginal values. The use of submodular functions is wide-spread: from information gathering \cite{krause2011submodularity} and image segmentation \cite{boykov2001interactive} to document summarization \cite{lin2011class}. Rockafellar and Uryasev \cite{rockafellar2000optimization} introduce a relationship between a  submodular function and the Conditional-Value-at-Risk (CVaR). CVaR is a risk metric that is commonly employed in stochastic optimization in finance and stock portfolio optimization.
%They then propose an auxiliary function by which the CVaR can be maximized. 
Another popular measure of risk is the Value-at-Risk (VaR) \cite{morgan1997creditmetrics}, which is commonly used to formulate the chance-constrained optimization problems.
In \cite{yang2017algorithm}, and \cite{yang2018algorithm}, the authors study the chance-constrained optimization problem while also considering risk in the multi-robot assignment, and then extend it to a knapsack formulation. In a comparison between the VaR and CVaR, Majumdar and Pavone \cite{majumdar2020should} propose that the CVaR is a better measure of risk for robotics, especially when the risk can cause a huge loss. In \cite{zhou2018approximation} and \cite{zhou2020risk}, a greedy algorithm for maximizing the CVaR is proposed.  

Building on the work by \cite{zhou2018approximation,zhou2020risk}, in this paper, %we leverage this prior work on choosing an unordered set to develop the first risk-aware optimization for choosing an ordered tour. {Our major contribution is the
we develop a polynomial-time algorithm for approximating a solution to the stochastic TSP. Our method differs from the previous approaches due to the presence of uncertainty in the tour cost, making traditional path-planning algorithms fail. The framework presented in \cite{zhou2018approximation,zhou2020risk} is effective only for one-stage planning (selecting a path amongst a set of candidate paths). In this paper, we present a multi-stage planner that finds a route taking the stochastic aspect into account. To achieve this, we propose an objective function that balances  risk and reward for a tour and prove that this function is submodular. %  whereas the algorithm developed in this paper finds a balance between two opposing functions formed from the rewards and costs. Our algorithm plans an entire TSP tour, which is fundamentally more challenging than selecting a path amongst a set of candidate paths for the vehicle assignment in \cite{zhou2018approximation,zhou2020risk
%We can say 15-16 were not designed for finding tours but just selecting from amongst a set of candidate paths. Here, we are *planning* for a path so the algorithm needs to take that into account. We can also say that another novel contribution is this function that balancesand showing that it is submodular.
 The method in \cite{jawaid2013maximum} addresses the deterministic TSP with a reward-cost trade-off, while our work is focused  on a stochastic version where the uncertainty in reward and cost is considered. The algorithm from \cite{jawaid2013maximum} can be viewed as a special case of our algorithm, with $\alpha=1$ and the subsequent risk ignored.
%We determine a single solution set to maximize the CVaR using a monotone submodular function \cite{zhou2018approximation}. The optimization problem is then framed over a matroid constrained system.

\noindent \textbf{Contributions}:  The main contributions of this paper are:
\begin{itemize}%Risk-Aware Multi Objective%
    \item \textcolor{black}{We present a risk-aware TSP with a stochastic objective that balances risk and reward for planing a TSP tour (Problem~\ref{prob:risk-tsp}). 
    \item We show the objective is submodular (Lemma~\ref{lemma:sub and monotone}) and }propose a greedy algorithm (\texttt{RAGA}) to find tours that maximize the CVaR of a stochastic objective (Algorithm~\ref{alg:riskg}).
    \item We prove that the solution obtained by \texttt{RAGA} is within a constant approximation factor of the optimal and an additive term proportional to the optimal solution (Theorem~\ref{thm:approximation}) and prove that \texttt{RAGA} has a polynomial run-time (Theorem~\ref{thm:complexity}). 
    \item We evaluate the performance of the algorithm through extensive simulations (Section~\ref{sec: results}).
\end{itemize}

% \indent \textbf{Organization}: The rest of the paper is organized as follows. In Sec. \ref{sec: prelim}, the necessary background materials on  sets, function properties, VaR, and CVaR is given. In Sec. \ref{sec: prob_form}, we describe risk-aware  TSP and the multi-objective optimization problem. Subsequently, we present our main algorithm \texttt{RAGA} and then derive its performance in Sec. \ref{sec: algo_&_analysis}. In Sec. \ref{sec: results}, we show simulation results of \texttt{RAGA}. Finally, in Sec. \ref{sec: conc}, we conclude and discuss future work.
% \pb{Check if the revised intro is alright. Otherwise, intro_new can be replaced by intro_old.}

\section{Preliminaries}\label{sec: prelim}

We first introduce the conventions and notations used in this paper. Calligraphic capital letters denote sets (e.g. $\mathcal{A}$). $2^{\mathcal{A}}$ denotes the power set of $\mathcal{A}$ and $|\mathcal{A}|$ represents its cardinality. Given a set $\mathcal{B}$, $\mathcal{A \setminus B}$ denotes set difference. Let $x$ be a random variable, then $\mathbb{E}[x]$ represents the expectation of the random variable $x$, and $\mathbb{P[\cdot]}$ denotes its probability. 

\subsection{Set and Function Properties}

Optimization problems generally work over a set system $(\mathcal{X, Y})$ where $\mathcal{X}$ is the base set and $\mathcal{Y} \subseteq 2^{\mathcal{X}}$. A reward/cost function $f : \mathcal{Y} \rightarrow \mathbb{R}$ is then either maximized or minimized. 
% \begin{definition}
% (Independence System): A set system that is closed under subsets (i.e., if $\mathcal{A} \in \mathcal{Y}$, and $\mathcal{B} \subseteq \mathcal{A}$, then $\mathcal{B} \in \mathcal{Y}$). All sets in such $\mathcal{Y}$ are defined as \textit{independent sets}. \textbf{Bases} are defined as the set of maximal independent sets (i.e., all $\mathcal{A}$ such that $\mathcal{A} \in \mathcal{Y}$ and $\mathcal{A} \cup \{x\} \notin \mathcal{Y}, ~\forall x \in \mathcal{X} \setminus \mathcal{A}$). %\lz{all A?}
% \end{definition}

% \begin{definition}
% (p-System): Given an independence system $\mathcal{(X, Y)}$. For all $\mathcal{A} \subseteq \mathcal{X}$, let:
% \begin{eqnarray*} 
%     \mathsf{U}\mathcal{(A)} = \max_{\mathcal{B}:\: \mathcal{B}\: is\: the\: basis\: of\: \mathcal{A}} \;|\mathcal{B}|\\
%     \mathsf{L}\mathcal{(A)} = \min_{\mathcal{B}:\: \mathcal{B}\: is\: the\: basis\: of\: \mathcal{A}} \;|\mathcal{B}|
% \end{eqnarray*}
% be the sizes of the maximum and minimum cardinality of the bases of $\mathcal{A}$. Then the pair $\mathcal{(X,Y)}$ is a \textit{p-system} if:
% \begin{equation}
%     \mathsf{U}(\mathcal{A}) \leq p\mathsf{L}(\mathcal{A}).
% \end{equation}
% \end{definition}
\begin{definition}
\label{def:inc}
(Monotonically Increasing):  A set function $f : \mathcal{Y} \rightarrow \mathbb{R}$ is said to be monotonically increasing if and only if for any set $\mathcal{S'} \subseteq \mathcal{S} \in 2^{\mathcal{X}}, ~f(\mathcal{S'}) < f(\mathcal{S})$.
\end{definition}
% \begin{definition}
% (Normalized Function): A function $f : \mathcal{Y} \rightarrow \mathbb{R}$ is normalized if $f(\phi) = 0$, where $\phi$ is the null set.
% \end{definition}
\begin{definition}
\label{def:sub}
(Submodularity): A function $f:2^{\mathcal{X}} \rightarrow \mathbb{R}$ is submodular if and only if\\
%\begin{multline}
$f(\mathcal{S}) + f(\mathcal{T}) \geq f(\mathcal{S} \cup \mathcal{T}) + f(\mathcal{S} \cap \mathcal{T}), ~\forall \mathcal{S}, \mathcal{T} \in 2^{\mathcal{X}}$.
%\end{multline}
% We can also say that $f$ is submodular if:
% \begin{equation}
% f(\mathcal{S}) + f(\{x\}) \geq f(\mathcal{S} \cup \{x\}) + f(\mathcal{S} \cap \{x\}).
% \end{equation}
% Submodular functions thus have diminishing marginal returns. More formally, let us define $\Delta_{\mathcal{S}}x = f(\mathcal{S} \cup \{x\})-f(\mathcal{S})$. Then, $f$ is a submodular function if,
% \begin{equation}
%     \Delta_{\mathcal{S'}}x \geq \Delta_{\mathcal{S}}x, \; \forall \mathcal{S'}\subseteq \mathcal{S} \subseteq 2^{\mathcal{X}}
% \end{equation}
\end{definition}
\begin{definition}
(Matroid): An independence set system $\mathcal{(X,Y)}$ is called a matroid if for any sets $\mathcal{S, P} \in 2^{\mathcal{X}}$ and $|\mathcal{P}| \leq |\mathcal{S}|$, it must hold that there exists an element $s \in \mathcal{S} \setminus \mathcal{P}$ such that $\mathcal{P} \cup \{s\} \in \mathcal{Y}$.
\end{definition}
% \begin{definition}
% (Partition Matroid): A partition matroid is a matroid $\mathcal{(X, Y)}$ in which, for a positive integer $n$, there exists disjoint sets $\mathcal{X}_{1}, \ldots \mathcal{X}_{n}$, and positive integers $\mathit{k}_{1}, \dots, \mathit{k}_{n}$ such that $\mathcal{X} \equiv \mathcal{X}_{1} \ldots \mathcal{X}_{n}$, and $\mathcal{Y} = \{\mathcal{S} : \mathcal{S} \subseteq \mathcal{X}, |\mathcal{S} \cap \mathcal{X}_{i}| \leq k_{i}, ~\forall i=1, \ldots n \}$.
% \end{definition}
\begin{definition}
(Curvature): Curvature is used as a measure of the degree of submodularity of a function $f$. Consider the matroid pair $(\mathcal{X, Y})$, and a function $f: 2^{\mathcal{X}} \rightarrow \mathbb{R}$, such that for any element $s \in \mathcal{X}, f(\{s\}) \neq 0$. The curvature $k,0 \leq k \leq 1$ is then defined as:
\begin{equation}
    k = 1 - \min_{s \in \mathcal{S}, \mathcal{S} \in \mathcal{Y}} \: \frac{f(\mathcal{S}) - f(\mathcal{S}\setminus \{s\})}{f(\{s\})}.
\end{equation}
%Note that by the above definition $0 \leq k \leq 1$. 

% Equivalently the curvature can be found by solving:
% \begin{equation}
%     \min_k \; \Delta_{S}(s) \geq (1-k)f(\{s\})
% \end{equation}
% \lz{where is minimum and maximum? the definition is not clear, please refer to my WAFR paper.}
\end{definition}

\subsection{Travelling Salesperson Problem}

% \begin{definition}
%     (Finite Graph): A finite graph is an ordered pair $G(\mathcal{V, E})$, where $\mathcal{V}$ is a set of finite number of vertices, and each element of $\mathcal{E}$ is a 2-element subset of $\mathcal{V}$, denoting the edge between these two vertices.
% \end{definition}

% \begin{definition}
%     (Weighted Graph): A weighted graph is a finite graph where each edge is associated with a specific weight (e.g., reward, length, cost, etc).
% \end{definition}

% \begin{definition}
%     (Complete Graph): A complete graph is a finite graph in which every node has an edge to every other node in the graph
% \end{definition}

% \begin{definition}
%     (Walk): A sequence $\mathit{v_{0}, e_{0}, v_{1}, e_{1}, ...}$ where $v_{i} \in \mathcal{V}$ and $e_{i} \in \mathcal{E}, ~\forall i$.
% \end{definition}

% \begin{definition}
%     (Tour): A walk with no repeated edges is called a tour.
% \end{definition}

% \begin{definition}
%     (Cycle): A cycle is a walk with no repeated edges, in which the only recurring vertices are the first and last vertices.
% \end{definition}

% \begin{definition}
%     (Hamiltonian Cycle): A Hamiltonian cycle is a sequence $\mathit{v_{0}, e_{0}, v_{1}, e_{1}, ...}$ where $v_{i} \in \mathcal{V}$ and $e_{i} \in \mathcal{E}, ~\forall i$ with no repeated edges, in which the only recurring vertices are the first and last vertices.
% \end{definition}

\begin{definition}
(TSP): Given a complete graph $G(\mathcal{V, E})$, the objective of the TSP is to find a minimum cost (maximum reward) Hamiltonian cycle. 
\end{definition}
In this paper, we consider the symmetric undirected TSP, where each edge has a reward and cost associated with it.

% Note that the symmetric TSP is an intersection of two matroids: 
% (i)A partition matroid with the degree of each vertex $\leq$ 2;
% and (ii) A 1-graphic matroid containing the set of edges that form a forest with at most one simple cycle.

% \begin{theorem}
% \textup{(Jenkyns, \cite{jenkyns1979greedy})}. On a grpah with n vertices, the Symmetric TSP (STSP) is a p-system with $p = 2 - \lfloor {\frac{n+1}{2}}\rfloor^{-1}  < 2 $ .
% \end{theorem}

% \subsection{Greedy Approximation Algorithm}
% The greedy algorithm is an approximation algorithm, where at each instant, the element optimizing a cost function is selected. In the algorithm we propose, we maximize the monotone submodular cost function $f$.
% For any matroid system, the greedy approximation algorithm gives a $\frac{1}{2}$ approximation. If we know the curvature $\mathit{k}$ of the submodular set function $f$  then we will have a $\frac{1}{2+\mathit{k}}$ approximation (Lemma 2.15 \cite{jawaid2013maximum})

\subsection{Measure of Risk}
Let $f(\mathcal{S}, y)$ denote a utility function with solution set $\mathcal{S}$ and noise $y$. As a result of $y$, the value of $f(\mathcal{S},y)$ is a random variable for every $\mathcal{S}$. 
\begin{definition}
    (Value at Risk): The Value at Risk (VaR) is defined as:
\begin{equation}
    \emph{VaR}_{\alpha}(\mathcal{S}) = \min_{\tau \in \mathbb{R}} \; \{\mathbb{P}[f(\mathcal{S}, y) \leq \tau] \geq \alpha \}, \; \alpha \in (0,1],
\end{equation}
\textcolor{black}{where $\alpha$ is the user-defined risk-level. A higher value of $\alpha$ corresponds to the choice of a higher risk level}.
% , which implies a larger value variance/uncertainty induced on the set $\mathcal{S}$
% The VaR is the left-endpoint of the non-empty interval of the solution to 
% \begin{equation}
% \int_{f(\mathcal{S}, y) \leq \tau} \: p(y) \,dy = \alpha
% \end{equation}
\end{definition}

\begin{definition}
    (Conditional Value at Risk): The Conditional-Value-at-Risk (CVaR) is defined as
\begin{equation}
    \emph{CVaR}_{\alpha}(\mathcal{S}) = \mathbb{E}_{y} [f(\mathcal{S},y) \:|\: f(\mathcal{S},y) \leq \emph{VaR}_{\alpha}(\mathcal{S})].
\end{equation}
Maximizing the value of $\text{CVaR}_{\alpha}(\mathcal{S})$ is equivalent to maximizing the auxiliary function $H(\mathcal{S}, \tau)$ (Theorem 2, \cite{rockafellar2000optimization}):
\begin{equation}
    H(\mathcal{S}, \tau) = \tau - \frac{1}{\alpha}\mathbb{E}_y[\: (\tau - f(\mathcal{S}, y))^{+} \:],
\end{equation}
where $[t]^{+} = t, ~\forall t \geq 0$ and $0$ when $t < 0$.
\end{definition}

% Fig. \ref{fig:cvar} shows an example of the VaR and the CVaR for a particular distribution of the utility function $f(\mathcal{S}, y)$

\begin{lemma}
\label{lemma:properties}
\textup{(Lemma 1, \cite{zhou2018approximation})} If $f(\mathcal{S},y)$ is monotone increasing, submodular and normalized in set $\mathcal{S}$ for any realization of $y$, then the auxiliary function $H(\mathcal{S}, \tau)$ is monotone increasing and submodular but not necessarily normalized \footnote{\textcolor{black}{The function $f(\mathcal{S}, y)$ is normalized in $\mathcal{S}$ if and only if $f(\emptyset, y) = 0$.}} in set $\mathcal{S}$ for any given $\tau$.
\end{lemma} 

\begin{lemma}
\textup{(Lemma 2, \cite{zhou2018approximation})} The auxiliary function $H(\mathcal{S}, \tau)$ is concave in $\tau, \forall\: \mathcal{S}$.
\end{lemma}

\section{Problem Formulation} \label{sec: prob_form}
In this section, we first discuss a risk-aware TSP and then formulate the problem as a  stochastic optimization problem by using CVaR.

% \begin{figure}
% 	\centering
% 	\includegraphics[width=\linewidth]{cvar.png}
% 	\caption{An illustration showing the VaR and the CVaR.}\label{fig:cvar} \lz{this CVaR should be redrawn, cannot be directly adopted.}  
% \end{figure}

\subsection{Risk-Aware TSP}
In order to motivate our formulation, consider the scenario of monitoring active volcanoes using a robot (say aerial robot) on an island as shown in Fig. \ref{fig:map_Hawaii}, where the red-colored patches represent active volcanic regions.  The important sites ($\mathcal{V}$) that the robot needs to visit are shown in Fig. \ref{fig:key_points}. \textcolor{black}{These sites are strategic positions from which it is possible to observe the volcanic situation from a safe distance.} The robot travels between these monitoring sites and receives a reward based on the information gathered while traversing this tour. In this work, we do not constrain vehicle motion in terms of distance or time.  While traveling directly above the volcano, the robot faces a higher chance of failure (due to volcanic activity) but can gather more information (a higher reward), while traveling along a shorter path (less cost). On the other hand, while traveling around the volcano, the robot has a lower risk but must travel a longer distance (more cost) while also receiving a lower reward. Fig. \ref{fig:low_cost_low_reward} and Fig. \ref{fig:high_cost_high_reward} show the paths that could be adopted based on the risk level specified for the robot. Our objective is to find a suitable tour for a single robot while considering the risk threshold and the trade-off between path cost and reward required. 

\begin{figure}
    \centering
    \includegraphics[width=0.8\linewidth]{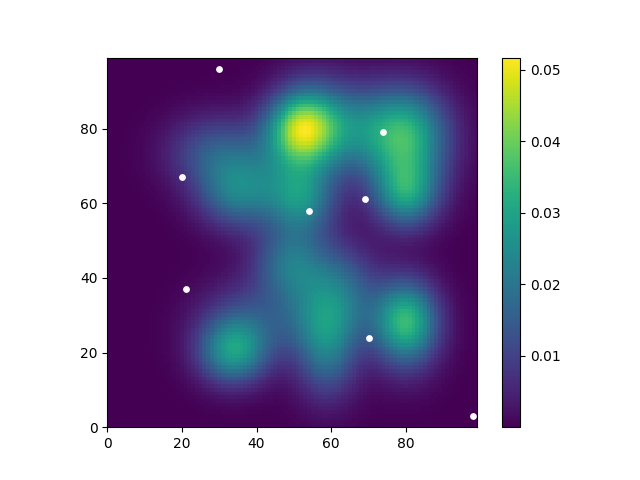}
    \caption{Map $M$ shows the information distribution in the environment $E$. The right bar shows the density degree of the information.}
    \label{fig:info_map}
\end{figure}

The monitoring task is modeled as a risk-aware TSP on a graph $G(\mathcal{V}, \mathcal{E})$ of an environment ${E}$ with $|\mathcal{V}|$ sites of interest. The notation $\mathcal{E}$ represents the set of edges connecting the vertices. A representative information density map ${M}$ of the environment ${E}$ is shown in Fig. \ref{fig:info_map}. The robot has sensors with a sensing range $R$. As it travels along the tour, the robot receives a reward based on the amount of information it collects and a penalty proportional to the tour's cost. We assume that both the reward and cost are random variables, with the reward being positive, and the cost of every edge having an upper-bound of $C$. \textcolor{black}{We use $r(\mathcal{S}, y_{r})$ and $c(\mathcal{S}, y_{c})$ to denote the reward and cost for a set of edges $\mathcal{S}$, where $r(\mathcal{S})$ and $c(\mathcal{S})$ are the sum of rewards of all points observed and the costs incurred when travelling along the edges in set $\mathcal{S}$ respectively, and  $y_{r}$ and $y_{c}$ are the respective noises induced.}

As we want to minimize cost and maximize reward simultaneously, our utility function $f(\mathcal{S},y)$ is a combination of these two terms, with a weighting factor deciding the priority we place on the cost over the reward. We define $f(\mathcal{S},y)$ as
\begin{equation}
    f(\mathcal{S}, y) = (1-\beta)\:r(\mathcal{S}, y_{r}) +  \beta\:(|\mathcal{S}|C-c(\mathcal{S}, y_{c})),
\end{equation}
where $\beta \in [0,1]$ is the weighting parameter. When $\beta = 0$, we ignore the cost incurred and consider the rewards received only. When $\beta = 1$, we ignore the rewards and are wary of only the cost penalized (classical TSP). Note that we directly incorporate the cost-reward trade-off into the utility function.

Let us define $r(\mathcal{S}, y_{r})$ as $f_{r}(\mathcal{S}, y_{r})$ and $|\mathcal{S}|C-c(\mathcal{S}, y_{c})$ as $f_{c}(\mathcal{S}, y_{c})$. 

% Formally, let us define the mean and variance of the reward received while travelling along the set of edges in $\mathcal{S}$ as $(r(\mathcal{S}), y_{r})$, and the corresponding mean and variance in costs as $(c(\mathcal{S}), y_{c})$. 

%One thing to note would be that the magnitudes of the rewards and cost would differ. As a result, $\beta$ could be biased to either  0 or 1. To overcome this, we first normalize both the reward and costs and then re-scale them to a common factor. 

\begin{lemma}
The utility function $f(\mathcal{S}, y)$ is both submodular and monotone increasing in $\mathcal{S}$.
\label{lemma:sub and monotone}
\end{lemma}

%%%%%%%%%%%%%%NEW VERSION TEXT

\proof \textcolor{black}{Let $e$ be the edge to be added to the tour, where $e \in \mathcal{E} \setminus \mathcal{S}$ and $\mathcal{S}$ is set of edges selected so far. Let us consider the two parts of $f(\mathcal{S}, y)$ separately. \\
\paragraph{$f_{r}(\mathcal{S},y)$}
    \begin{itemize}[leftmargin=*]
        \item As the rewards are sampled from a truncated Gaussian, with a lower bound of 0, they are thus always positive. Therefore, $f_{r}(\mathcal{S}, y)$ is always monotone increasing in $\mathcal{S}$,
        \begin{equation}
            f_{r}(\mathcal{S}) \leq f_{r}(\mathcal{S}\cup\{e\}).
        \end{equation}
        % \lz{better to write down the monotonicity equation}
        \item While calculating the total reward, we add the rewards obtained from all \textit{distinct} points observed while traversing this tour. Consider the current set of edges $\mathcal{S}$ and a new edge $e$. If the new edge $e$ and the edges $\mathcal{S}$ have any overlap in the sensed regions, the total reward received from traversing $\mathcal{S}$ and $e$ successively will be less than the sum of rewards of traversing $\mathcal{S}$ and $e$ individually. Therefore,
        \begin{equation*}
            f_{r}(\mathcal{S} \cup \{e\})  \leq f_{r}(\mathcal{S}) + f_{r}(\{e\}).
        \end{equation*}
        Also, since $e \in \mathcal{E} \setminus \mathcal{S}$, $f_{r}(\mathcal{S} \cap \{e\}) =0$. Then we have, 
        \begin{equation}
            f_{r}(\mathcal{S} \cup \{e\}) + f_{r}(\mathcal{S} \cap \{e\}) \leq f_{r}(\mathcal{S}) + f_{r}(\{e\}),
        \end{equation}
        and thus $f_{r}(\mathcal{S},y)$ is submodular in $\mathcal{S}$.
        %when the sensing radius ${R} > 0$ and modular only when ${R} = 0$. \lz{${R} = 0$? did we introduce how the robot's sensor model before?}
    \end{itemize}
    \vspace{2mm}
\paragraph{$f_{c}(\mathcal{S}, y)$}
    \begin{itemize}
        \item The cost of each edge is defined as a truncated Gaussian with a upper bound of $M_{c}$. For any set of edges in $\mathcal{S}$, the sum of costs will always be less than $|\mathcal{S}|M_{c}$, which means any sample (realization) of $f_{c}(\mathcal{S}, y)$ is positive. Therefore, $f_{c}(\mathcal{S}, y)$ is always monotone increasing in $\mathcal{S}$, i.e.,
        \begin{equation}
            f_{c}(\mathcal{S}) \leq f_{c}(\mathcal{S}\cup\{e\}).
        \end{equation}
         \item Consider again, the current set of edges $\mathcal{S}$ and the new edge $e$. The total cost of traversing $\mathcal{S}$ and $e$ is equal to the sum of costs of  traversing $\mathcal{S}$ and $e$ individually. Therefore,
        \begin{equation}
            f_{c}(\mathcal{S} \cup \{e\}) = f_{c}(\mathcal{S}) + f_{c}(\{e\}), 
        \end{equation}
        and thus $f_{c}(\mathcal{S}, y)$ is modular in $\mathcal{S}$.
    \end{itemize}
\paragraph{$f(\mathcal{S}, y)$} Since both $f_{r}(\mathcal{S}, y_{r})$ and $f_{c}(\mathcal{S}, y_{c})$ are monotone increasing in $\mathcal{S}$, $f(\mathcal{S}, y)$ as the summation of these two, is also monotone increasing in $\mathcal{S}$. Similarly, since $f_{r}(\mathcal{S}, y_{r})$ is submodular in $\mathcal{S}$ and $f_{c}(\mathcal{S}, y_{c})$ is modular in $\mathcal{S}$, $f(\mathcal{S}, y)$ as the summation of these two, is submodular in $\mathcal{S}$.  \hfill $\Box$ 
}

\subsection{Risk-Aware  Submodular Maximization}
Consider the set system $\mathcal{(E, I)}$, where $\mathcal{E}$ is the set of all edges in the graph $G(\mathcal{V, E})$. If $\mathcal{A}_{1}, \mathcal{A}_{2} \ldots \mathcal{A}_{n}$ each contains the edges forming $n$ Hamiltonian tours, then $\mathcal{I} := 2^{\mathcal{A}_{1}} \cup 2^{\mathcal{A}_{2}} \ldots \cup 2^{\mathcal{A}_{n}}$. We define our \textit{risk-aware  TSP} by maximizing  $\text{CVaR}_{\alpha}(\mathcal{S})$, where $\mathcal{S} \in \mathcal{I}$. We know that maximizing the $\text{CVaR}_{\alpha}(\mathcal{S})$ is equivalent to maximizing the auxiliary function ${H(\mathcal{S}, \tau)}$. Thus, we formally define the problem as:

\begin{problem} [Risk-aware  TSP]
\begin{equation}
    \max_{\mathcal{S} \in \mathcal{I}, ~\tau \in [0, \Gamma]} \tau - \frac{1}{\alpha} \mathbb{E}_y[ (\tau - f(\mathcal{S}, y))^{+}], 
\end{equation}
where $\Gamma$ is the upper bound on the value of $\tau$.
\label{prob:risk-tsp}
\end{problem}

\section{Algorithm And Analysis}\label{sec: algo_&_analysis}

\begin{algorithm}[h!]
\SetAlgoLined
\nonl \textbf{Input}: \\
\nonl \hskip0.5em Graph $G(\mathcal{V,E})$; Risk level $\alpha \in (0,1]$; Weighing factor $\beta \in [0,1]$; Upper bound $\Gamma \in \mathbb{R}^+$ on $\tau$; Searching factor $\gamma \in (0, \Gamma]$; Oracle function $\mathbb{O}$ that approximates $H(\mathcal{S}, \tau)$ as $\hat{H}(\mathcal{S}, \tau)$. \\
\nonl \textbf{Output}:\\
\nonl \hskip0.5em A Hamiltonian tour $S^{G}$ and its respective $\tau^{G}$.\\
\For {i = \{0, 1, 2, . . ., $\ceil*{\frac{\Gamma}{\gamma}}$\}}{\label{line:for_start}
  $\mathcal{S} = \emptyset$; $D = \mathbf{0}_{1 \times |\mathcal{V}|}$; \label{line:init} $\hat{H}_{\text{max}} = 0$; $\hat{H}_{\text{cur}} = 0$\\
   $\tau_{i} = i\,\gamma$ \\ \label{line:init_tau}
  \While {\label{line:while_start} $\mathcal{E} \neq \emptyset$ \textbf{and} $\mathcal{|S|} \leq |\mathcal{V}|$} {
    $e^\star = \text{argmax}_{{\forall e \in \mathcal{E} \setminus \mathcal{S}}} \; \hat{H}(\mathcal{S} \cup \{e\}, \tau_{i}) - \hat{H}(\mathcal{S}, \tau_{i})$\label{line:oracle}  \\ %\text{argmax}_{\forall e \in \mathcal{E} \setminus \mathcal{S}}
    $(u, v) \leftarrow \mathcal{V}(e^\star)$\\
    $\text{flag} = \texttt{False}$\\
    \If {$D[u]<2$ \textbf{and} $D[v]<2$}{\label{line:check_start}
        check subtour existence in $\mathcal{S}\cup \{e^{\star}\}$ by running DFS\label{line:DFS}\\
        \If {\text{no subtour present}}{
            $\text{flag} = \texttt{True}$
        }
    }\label{line:check_end}
    \If {$\text{flag} == \texttt{True}$}{\label{line:add_edge_start}
        $\mathcal{S} \leftarrow \mathcal{S} \cup \{e^\star\}$ \label{line:update}\\
        $D[u] += 1$; $D[v] += 1$\\
        $\hat{H}_{\text{cur}} = \hat{H}(\mathcal{S} \cup \{e^\star\}, \tau_{i})$
   } \label{line:add_edge_end}
   $\mathcal{E} \leftarrow \mathcal{E} \setminus \{e^\star\}$ \label{line:remove_e}
  }\label{line:while_end}
  \If {$\hat{H}_{\text{cur}} \geq \hat{H}_{\text{max}}$}{\label{line:best_tour_start}
    $\hat{H}_{\text{max}} = \hat{H}_{\text{cur}}$; $(\mathcal{S}^{G}, \tau^{G})  = (\mathcal{S}, \tau_{i})$
  }\label{line:best_tour_end}
  \If {$\hat{H}_{\text{cur}} < 0$}{\label{line:exit_start}
    \textbf{break}
  }\label{line:exit_end}
} \label{line:for_end}
 \caption{Risk-aware greedy algorithm (\texttt{RAGA})} \label{alg:riskg}
\end{algorithm}

In this section, we present a risk-aware greedy algorithm (\texttt{RAGA}) that extends the deterministic algorithm in \cite{jawaid2013maximum} for solving Problem~\ref{prob:risk-tsp}. We first explicitly introduce \texttt{RAGA}, then analyze its performance in terms of approximation bound and running efficiency.  
\begin{figure*}[th!]
\begin{multicols}{4}
    \includegraphics[width=\linewidth]{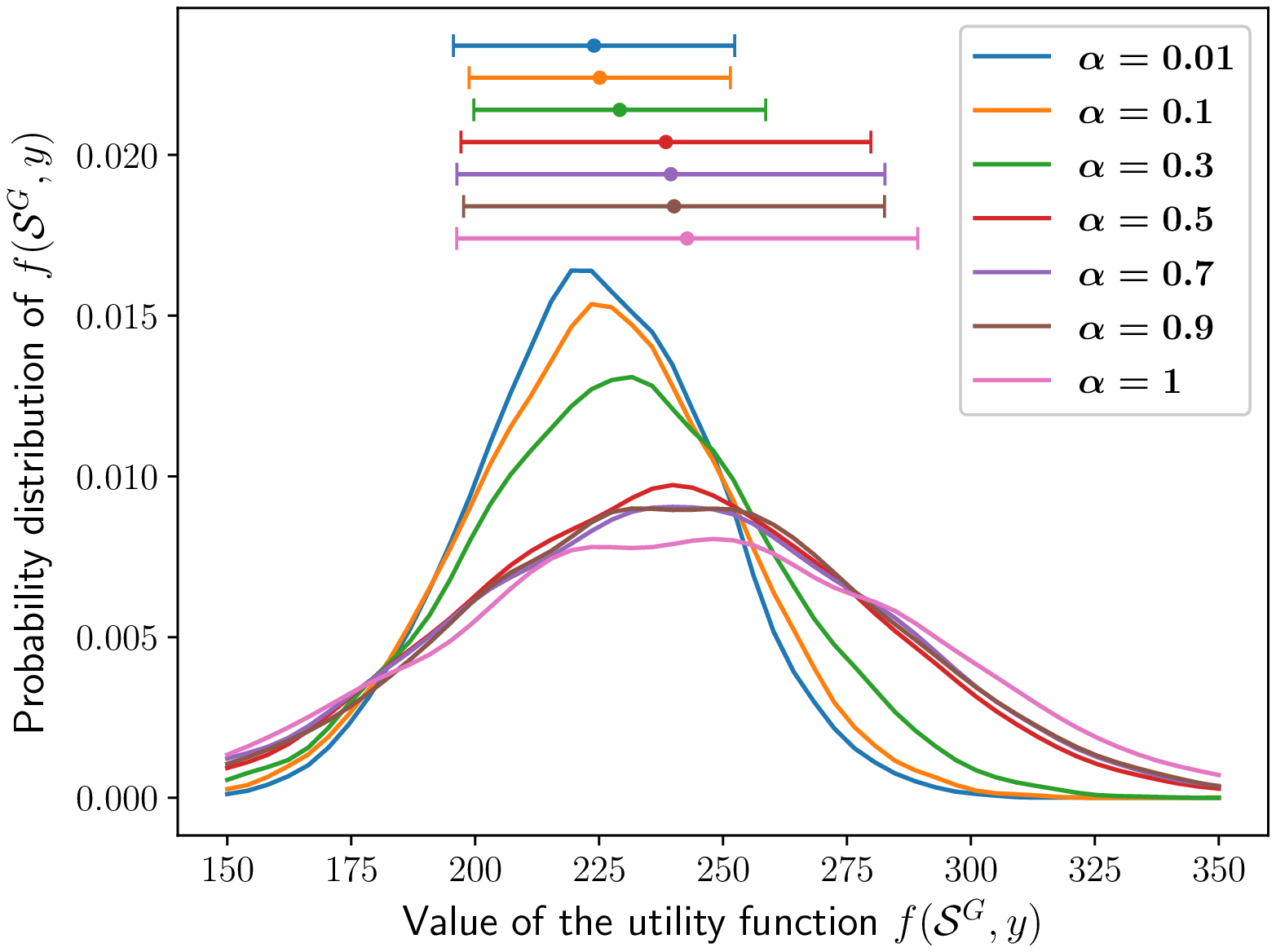}\subcaption{$\beta=0$}\label{fig:beta=0}
    \includegraphics[width=\linewidth]{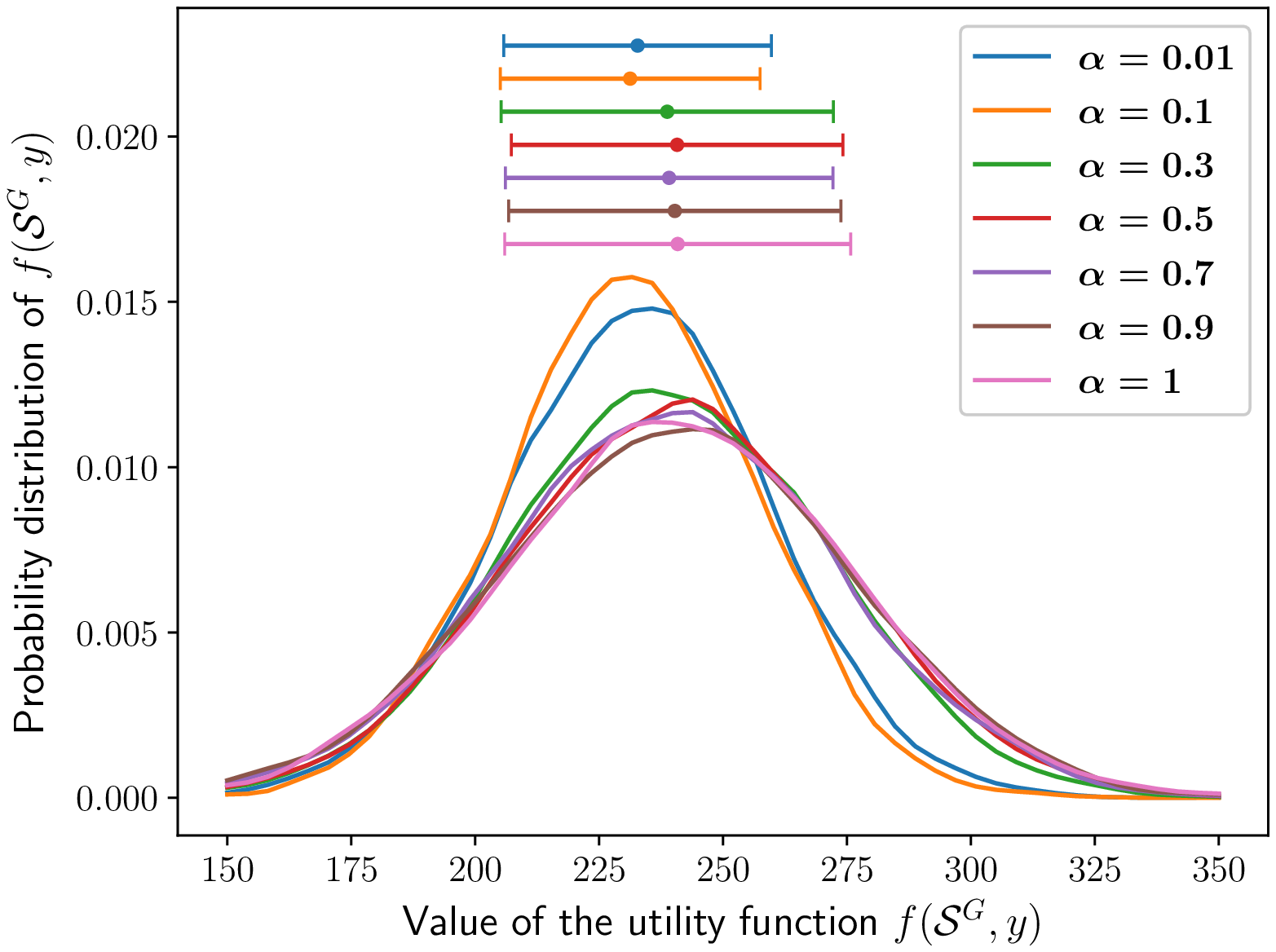}\subcaption{$\beta=0.4$}\label{fig:beta=0.4}
% \end{multicols}
% \begin{multicols}{2}
    \includegraphics[width=\linewidth]{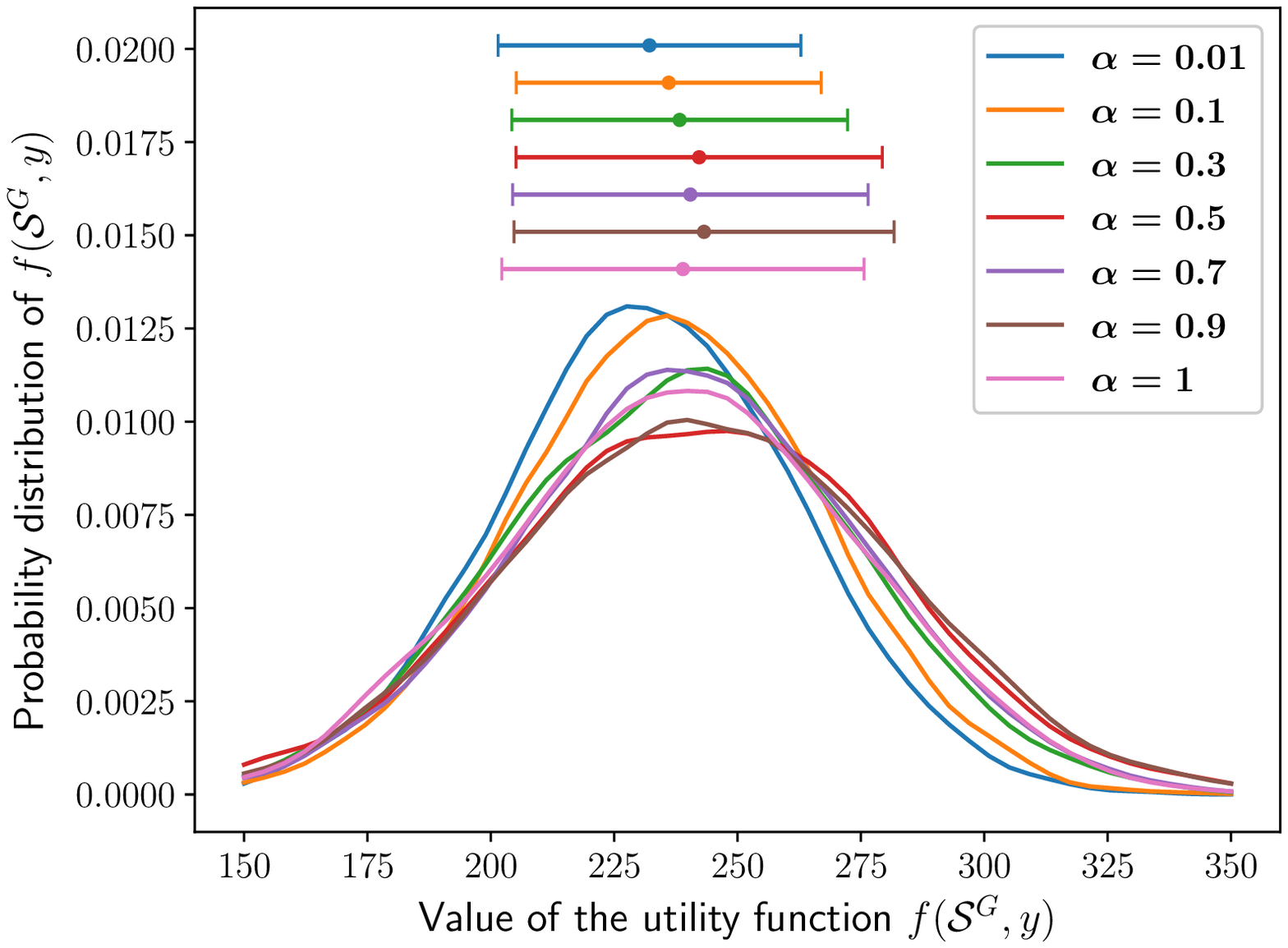}\subcaption{$\beta=0.7$}\label{fig:beta=0.7}
    \includegraphics[width=\linewidth]{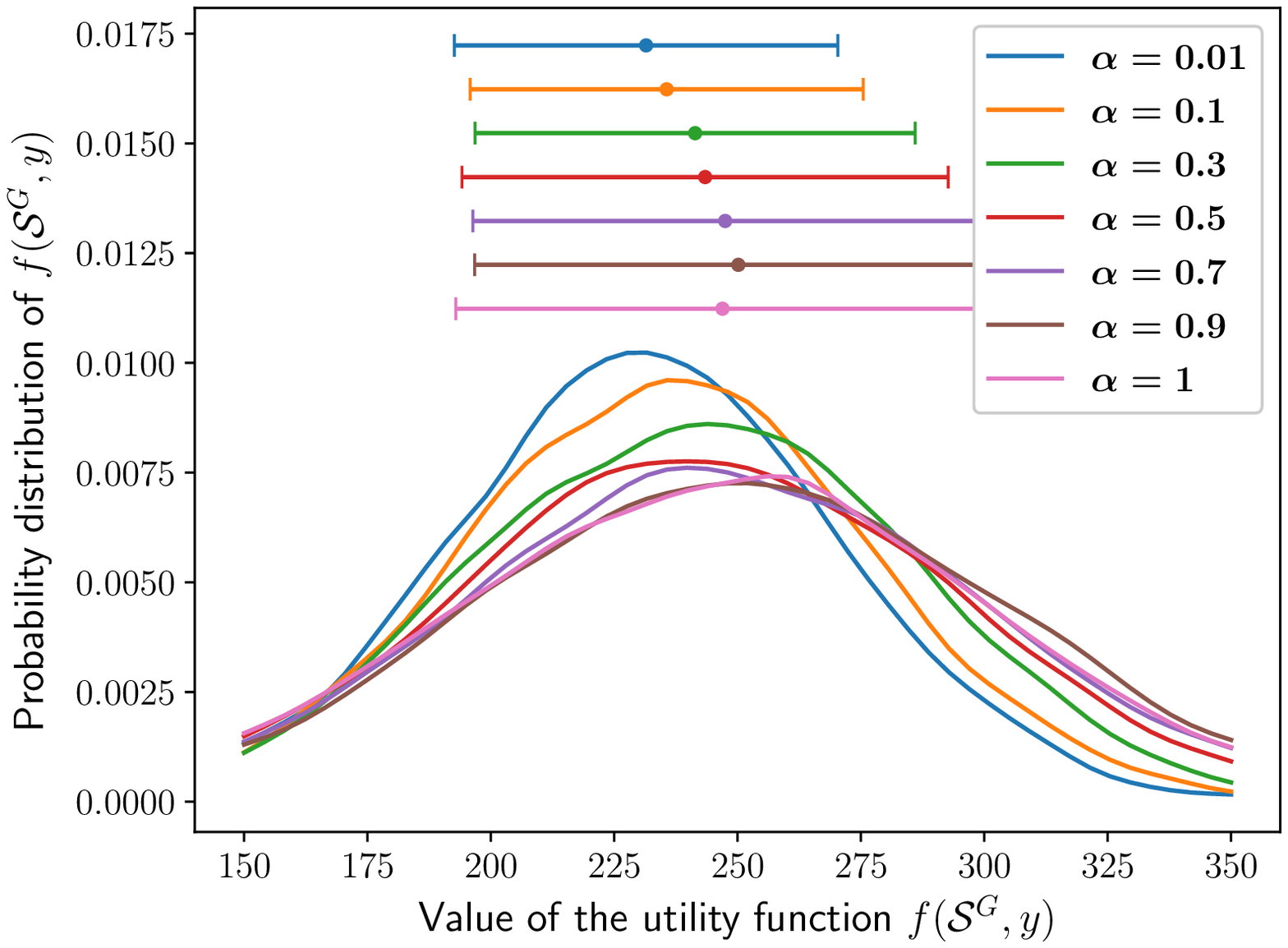}\subcaption{$\beta=1$}\label{fig:beta=1}
\end{multicols}
\vspace{-0.3cm}
\caption{Probability density function of $f(\mathcal{S}, y)$ at different values of $\alpha$, keeping $\beta$ fixed.}\label{fig:H for fixed beta} 
\end{figure*}

\begin{figure*}[th!]
\begin{multicols}{4}
    \includegraphics[width=\linewidth]{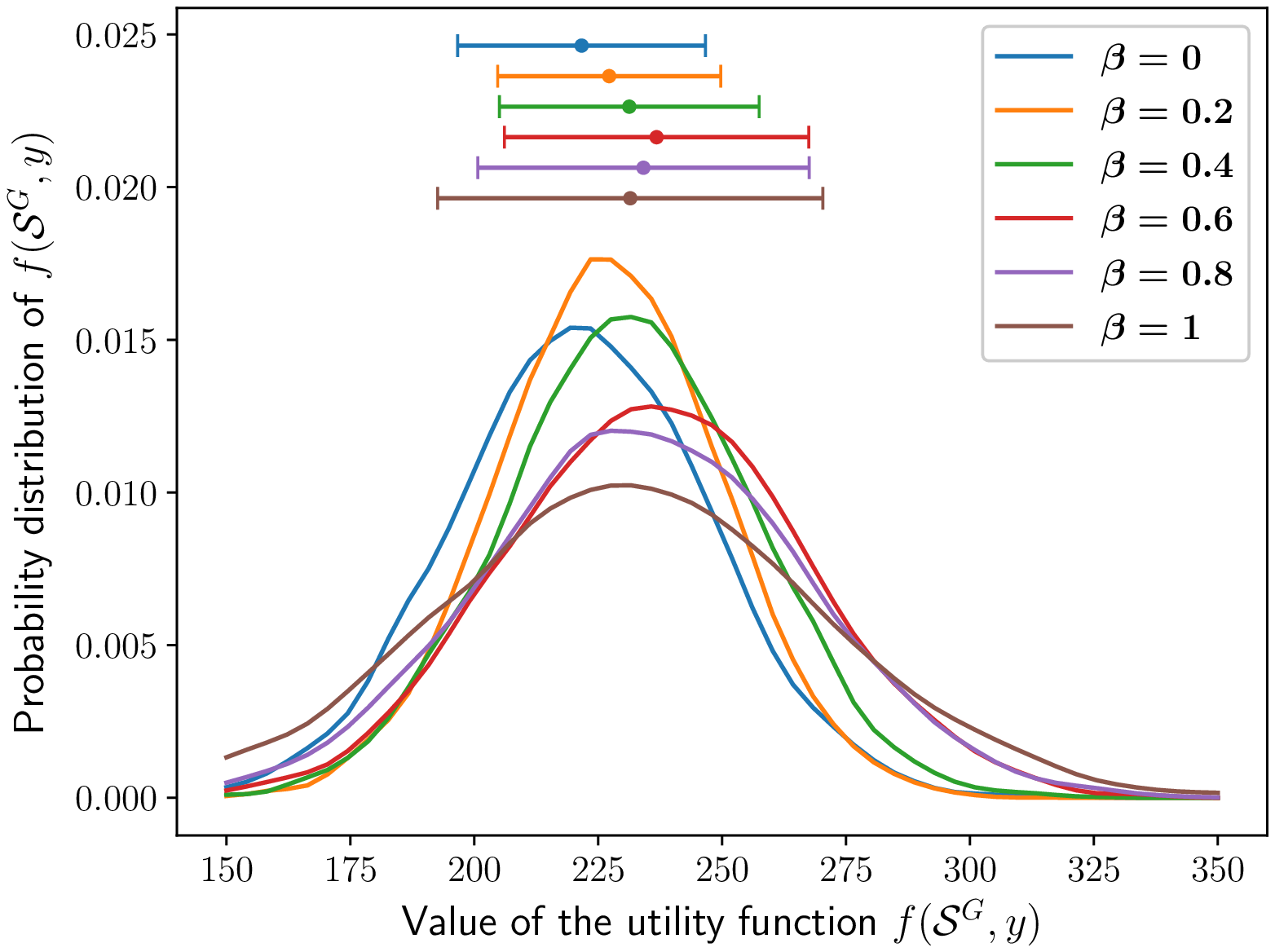}\subcaption{$\alpha=0.1$}\label{fig:alpha=0.1}\par
    \includegraphics[width=\linewidth]{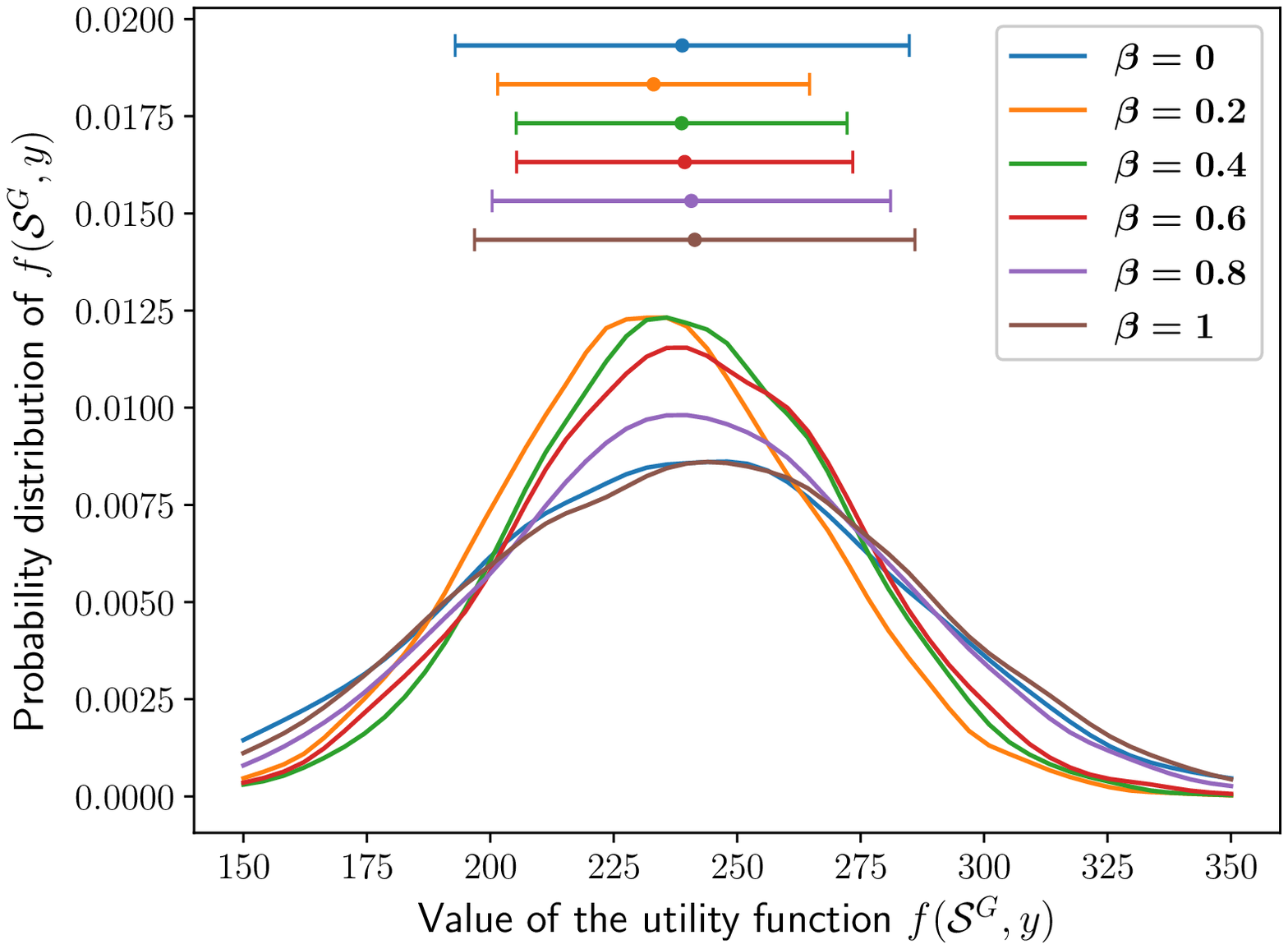}\subcaption{$\alpha=0.4$}\label{fig:alpha=0.4}\par 
% \end{multicols}
% \begin{multicols}{2}
    \includegraphics[width=\linewidth]{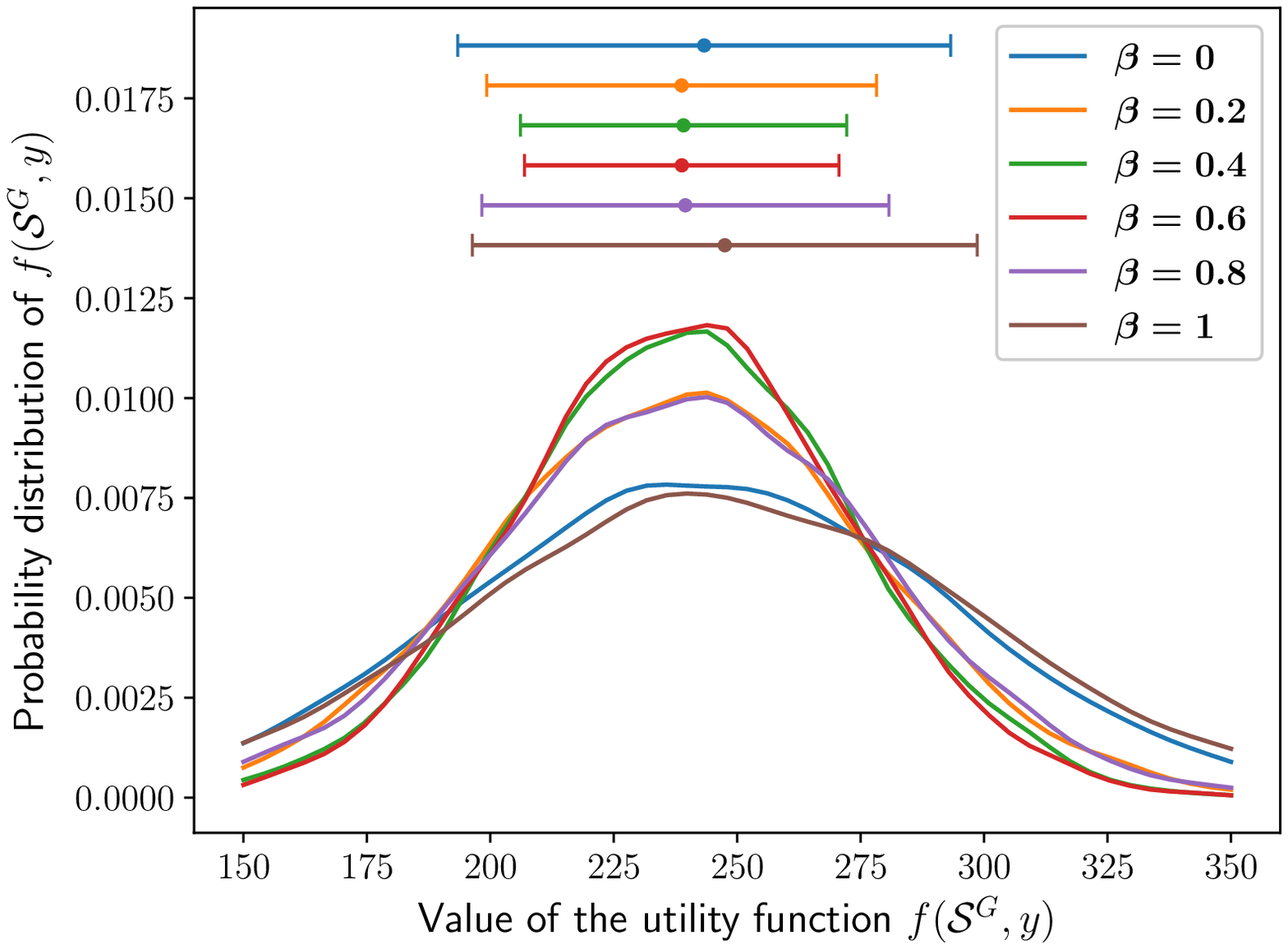}\subcaption{$\alpha=0.7$}\label{fig:alpha=0.7}\par
    \includegraphics[width=\linewidth]{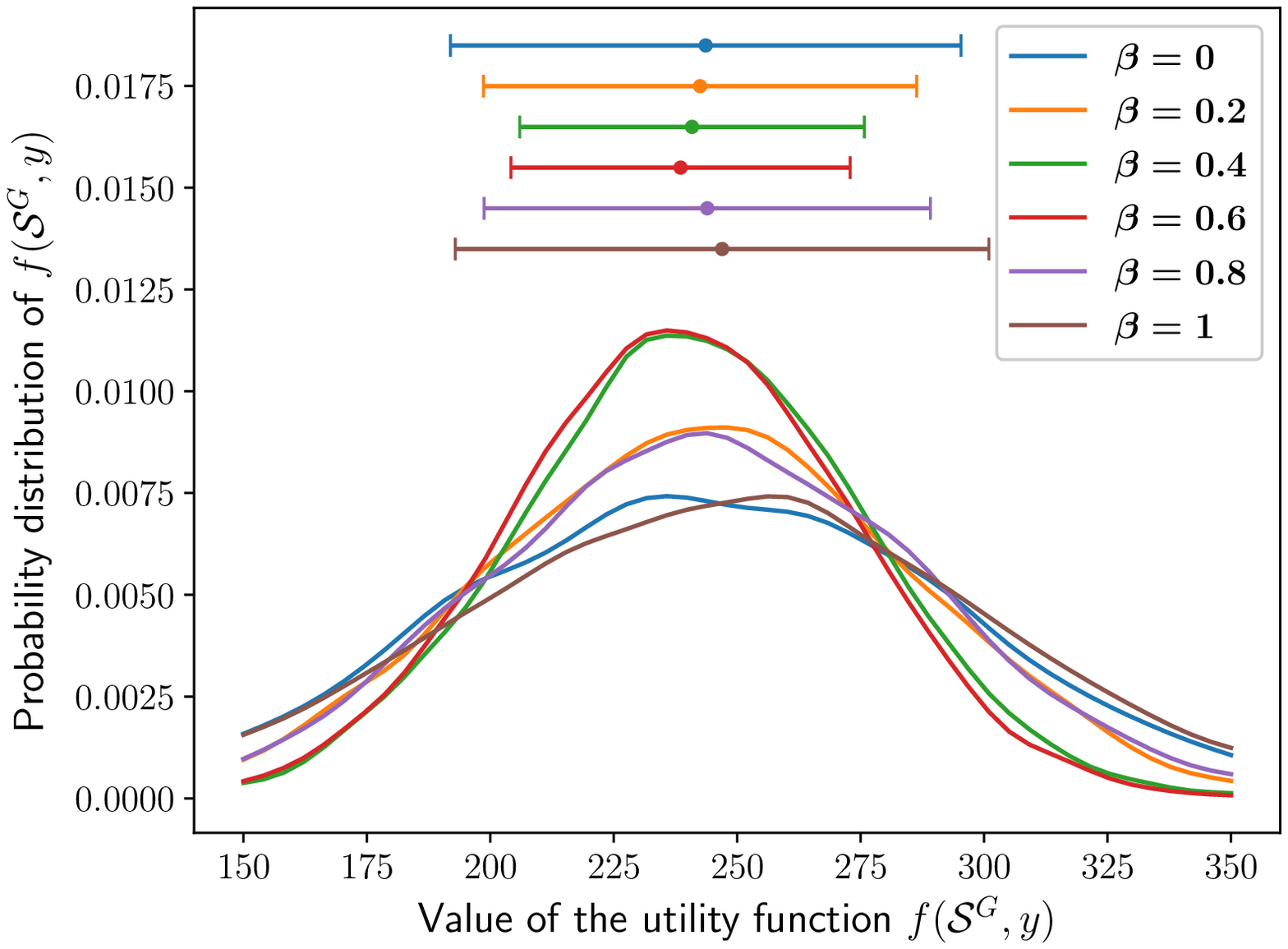}\subcaption{$\alpha=1$}\label{fig:alpha=1}\par
\end{multicols}
\vspace{-0.3cm}
\caption{Probability density function of $f(\mathcal{S}, y)$ at different values of $\beta$, keeping $\alpha$ fixed.}\label{fig:H for fixed alpha} 
\end{figure*}

\subsection{Algorithm}

\texttt{RAGA} has three main stages:\newline
%\begin{itemize}
    %\item 
    \noindent \textbf{(i) Initialization} (lines \ref{line:init} - \ref{line:init_tau})
    % \lz{it is better to add the line index here and after each sentence below. Same for other sections}:
    % We initialize the map ${M}$ and the respective graph $G\mathcal{(V,E)}$ on it. We initialize the discretization factor $\gamma$, risk-level $\alpha$, and importance factor $\beta$. We then set the maximum searching bound $\Gamma$ for $\tau$. 
    The decision set $\mathcal{S}$ (Hamiltonian cycle) is initialized as an empty set. A degree vector $D$ is set as $\mathbf{0}_{1 \times |\mathcal{V}|}$, which contains vertices' degrees in order. We use  variables $\hat{H}_{\text{max}}=0$ and $\hat{H}_{\text{cur}}=0$ store the maximal and the current values of $\hat{H}(\mathcal{S}, \tau)$ (line \ref{line:init}). \newline
    % In line \ref{line:init_tau} we initialize the value of $\tau$. 
    %\item 
    \noindent \textbf{(ii) Search for valid tour for every $\tau$} (lines \ref{line:while_start} - \ref{line:while_end}): For a specific value of $\tau$, we continue to add edges greedily until we have a complete tour or the edge set $\mathcal{E}$ is empty. For every edge $e \in \mathcal{E} \setminus \mathcal{S}$, we calculate the marginal gain of the auxiliary function, $\hat{H}(\mathcal{S} \cup \{e\}) - \hat{H}(\mathcal{S})$ using the oracle function $\mathbb{O}$, and choose the edge $e^{*}$ which maximizes the marginal gain (line \ref{line:oracle}). We then check if the selected edge $e^{*}$ forms (or could form) a valid tour with the elements in $\mathcal{S}$ (i.e., if $\mathcal{S} \cup \{e^{*}\} \in \mathcal{I}$) (lines \ref{line:check_start} - \ref{line:check_end}). Here, we first validate the degree constraint for each edge in $\mathcal{S} \cup \{e\}$, and then use the depth-first search (DFS) algorithm to iterate through the selected edges to check for subtours. %\footnote{A Hamiltonian cycle does not allow the existence of any subtour.}.
    If the new edge does not break the above constraints, we add the edge in $\mathcal{S}$ and update $D$ and $\hat{H}_{cur}$ (lines \ref{line:add_edge_start} - \ref{line:add_edge_end}). Finally, we remove $e^{*}$ from $\mathcal{E}$ in line \ref{line:remove_e}.\newline 
    %\item
    \noindent \textbf{(iii) Selecting best tour set ($\mathcal{S}^{G}, \tau^{G}$)} (lines \ref{line:best_tour_start} - \ref{line:best_tour_end}): For every tour we store the value of the auxiliary function $\hat{H}$ in $\hat{H}_{\text{cur}}$, and compare it to the best value $\hat{H}_{\text{max}}$. We  store the pair $(\mathcal{S}, \tau)$ whenever $\hat{H}_{\text{cur}}> \hat{H}_{\text{max}}$, and update $\hat{H}_{\text{max}}$. 
    % Thus we obtain the best value of the auxiliary function $\hat{H}_{\text{max}}$ and the corresponding $(\mathcal{S}, \tau)$ denoted as $(\mathcal{S}^{G}, \tau^{G})$. 
%\end{itemize}

Lines \ref{line:exit_start} - \ref{line:exit_end} show the condition of exiting the loop. As $H(\mathcal{S}, \tau)$ is concave in $\tau$, and we start from a non-negative value (as seen in Fig. \ref{fig:H_vs_tau}), it is certain that if $H(\mathcal{S}, \tau)$ becomes negative, it will continue to further decrease. This improves the runtime of \texttt{RAGA}.

\noindent \textbf{Designing the oracle function $\mathbb{O}$}: We use a sampling based oracle function to approximate $H(\mathcal{S}, \tau)$ as $\hat{H}(\mathcal{S}, \tau)$. The authors of \cite{ohsaka2017portfolio} have proved that if the number of samples $n_{s} = O(\frac{\Gamma^{2}}{\epsilon^{2}}\log\frac{1}{\delta})$, $\delta, \epsilon \in (0,1)$, the approximation for the value of CVaR (or equivalently, the auxiliary function $\hat{H}(\mathcal{S}, \tau)$) gives an error less than $\epsilon$ with a probability of at least $1-\delta$.

%\rb{Lifeng can you please verify this}

%\textcolor{blue}{Although our algorithm builds atop the one developed by the authors in \cite{jawaid2013maximum} by following a similar greedy approach for tour selection and \cite{zhou2018approximation} and \cite{zhou2020risk}, the algorithms are fundamentally different. The method in \cite{jawaid2013maximum} addresses the deterministic TSP with a reward-cost trade-off, while \texttt{RAGA} works on a stochastic problem where the uncertainty in reward and cost is considered. The algorithm from \cite{jawaid2013maximum} can be veiwed as a special case of \texttt{RAGA}, where $\alpha=1$ and the risk is ignored. Likewise, although the idea from \cite{zhou2018approximation} and \cite{zhou2020risk} form the foundation of our paper, the two algorithms differ in the formulation of the Oracle function $\mathbb{O}$.  }

\subsection{Performance Analysis}

\begin{theorem}
Let $\mathcal{S}^{G}$ and $\tau^G$ be the tour and the searching scalar selected by \texttt{RAGA} and let $\mathcal{S}^{\star}$ and $\tau^{\star}$ be the tour and the searching scalar selected the optimal solution. Then,
\begin{equation}
    H(\mathcal{S}^{G}, \tau^G)  \geq \frac{1}{2+\mathit{k}}(H(\mathcal{S}^{\star}, \tau^\star)-\gamma) + \frac{1+\mathit{k}}{2+\mathit{k}}\Gamma(1 - \frac{1}{\alpha}) - \epsilon
        \label{eqn:appro_bound}
\end{equation}
with a probability of at least $1-\delta$, when the number of samples $n_{s} = O(\frac{\Gamma^{2}}{\epsilon^{2}}\log\frac{1}{\delta}), \; \delta, \epsilon \in (0,1)$. $k$ is the curvature of $H(\mathcal{S}, \tau)$ with respect to $\mathcal{S}$.
\label{thm:approximation}
\end{theorem}

%\begin{proof}
\proof Note that $H(\mathcal{S}, \tau)$ is monotone increasing, submodular, but not normalized (Lemma \ref{lemma:properties}).  \cite[Theorem 6.1]{conforti1984submodular} and \cite[Theorem 2.10]{jawaid2013maximum} have shown that for a normalized submodular increasing function, the greedy algorithm gives an approximation of $\frac{1}{2+\mathit{k}}$ for maximizing it. Following this result, normalizing $H(\mathcal{S}, \tau)$ by $H(\mathcal{S}, \tau)-H(\emptyset, \tau)$, we get:
\begin{equation}
    \frac{H(\mathcal{S}^{G}, \tau) - H(\emptyset, \tau)}{H(\mathcal{S}^{\star}, \tau) - H(\emptyset, \tau)} \geq \frac{1}{2+\mathit{k}}
\end{equation}
where $H(\mathcal{S}^{\star}, \tau)$ is the optimal value of $H(\mathcal{S}, \tau)$ for any given value of $\tau$. Then following the proof of Theorem 1 in  \cite{zhou2020risk}, we have the bound approximation performance of \texttt{RAGA} in Equation~(\ref{eqn:appro_bound}).  $\hfill \Box$
% As $0 \leq \mathit{k} \leq 1, \: 0 < \alpha \leq 1$, and $H(\phi, \tau) = \tau - \frac{1}{\alpha}\tau$, we get:
% \begin{dmath}
%   H(\mathcal{S}^{G}, \tau)  \geq \frac{1}{2+\mathit{k}}H(\mathcal{S}^{*}, \tau) + \frac{1+\mathit{k}}{2+\mathit{k}}H(\phi, \tau)
%   \geq \frac{1}{2+\mathit{k}}H(\mathcal{S}^{*}, \tau) + \frac{1+\mathit{k}}{2+\mathit{k}}(\tau - \frac{1}{\alpha}\tau)
%   \geq \frac{1}{2+\mathit{k}}H(\mathcal{S}^{*}, \tau) + \frac{1+\mathit{k}}{2+\mathit{k}}(\Gamma - \frac{1}{\alpha}\Gamma)
% \end{dmath}
% {\pb for LZ: Some note on Pr(1-$\delta$)}
% %\end{proof}

\begin{theorem}
\texttt{RAGA} has a polynomial running time of $O(\lceil \frac{\Gamma}{\gamma} \rceil(|\mathcal{V}|^{3} (\mathcal{|V|}n_{p}+n_{s} + 2 \log |\mathcal{V}| + 1)))$.
\label{thm:complexity}
\end{theorem} 

%\begin{proof}
\proof First, the outer \textit{``for"} loop  (lines \ref{line:for_start}-\ref{line:for_end}) takes at most $\lceil \frac{\Gamma}{\gamma} \rceil$ time to search for $\tau$. Second, the inner \textit{``while"} loop (lines \ref{line:while_start}-\ref{line:while_end}) needs to check for edges $e \in \mathcal{E}$ to find the tour $\mathcal{S}$, thus running at most $|\mathcal{E}|$ times. 

Within the \textit{``while"} loop, \texttt{RAGA} finds the edge $e^{\star}$ with maximal marginal gain in line \ref{line:oracle}. To compute the marginal gain for an edge $e$, \texttt{RAGA} calls the oracle function $\mathbb{O}$. As it estimates $\hat{H}(\mathcal{S}, \tau_{i}), \forall e \in \mathcal{E} \setminus \mathcal{S}$, $\mathbb{O}$ is called at most $\mathcal{|E|}$ times, during one iteration of the \textit{``while"} loop in line \ref{line:while_start}. As \texttt{RAGA} needs to compute the marginal gains only when $\mathcal{S}$ is changed in line \ref{line:update}, and as $\mathcal{S}$ can have only $\mathcal{|V|}$ elements, it needs to perform a total of $|\mathcal{V}|$ recalculations. In addition, the edges need to be sorted every time a recalculation is performed, so that future calls to find the most beneficial element can be run in constant time without recomputing $\hat{H}(\mathcal{S}, \tau_{i})$.

For any edge $e$, the oracle $\mathbb{O}$ must compute the new mean reward and cost of $\mathcal{S} \cup \{e\}$. As only the distinct points in the environment are considered while calculating reward, the oracle must maintain an array of points that were observed while traveling along $\mathcal{S}$. If the maximum points observed when traveling along any edge in $\mathcal{E}$ is $n_{p}$, then the maximum number of points observed while traversing $\mathcal{S}$ would be $|\mathcal{V}|n_{p}$. Afterward, the oracle takes $n_{s}$ samples to compute $\hat{H}(\mathcal{S}, y)$. Thus the total runtime of the oracle is $\mathcal{|V|}n_{p}+n_{s}$. Therefore, the total time for computing gains and sorting would be $O(|\mathcal{V}|(|\mathcal{E}|(\mathcal{|V|}n_{p}+n_{s}) + |\mathcal{E}| \log |\mathcal{E}|))$.

Next, within the \textit{``while"} loop, \texttt{RAGA} checks if a the selected edge $e^{*}$ could form a valid tour with $\mathcal{S}$ (lines \ref{line:check_start}-\ref{line:check_end}). In particular, it ensures the degree of the two vertices of edge $e^\star$ is less than two (line \ref{line:check_start}). If both vertices have a degree less than two, it runs DFS to traverse the elements in $\mathcal{S}$ to check for subtours (line \ref{line:DFS}), which takes at most $|\mathcal{V}|$ time. As this is performed for every edge $e$ to be added to $\mathcal{S}$, in total, validating the selected edge runs in $O(|\mathcal{V}||\mathcal{E}|)$ time. 

Assuming that all other commands take constant time, \texttt{RAGA} has a runtime of $O(|\mathcal{V}|(|\mathcal{E}|(\mathcal{|V|}n_{p}+n_{s}) + |\mathcal{E}| \log |\mathcal{E}|) + |\mathcal{V}||\mathcal{E}|)$ for one iteration of the \textit{``for"} loop. Notably, for a complete graph, $|\mathcal{E}| = O(|\mathcal{V}|^{2})$. Considering that the \textit{``for"} loop has $\lceil \frac{\Gamma}{\gamma} \rceil$ iterations, the total runtime of \texttt{RAGA} becomes $O(\lceil \frac{\Gamma}{\gamma} \rceil(|\mathcal{V}|^{3} (\mathcal{|V|}n_{p}+n_{s} + 2 \log |\mathcal{V}| + 1)))$. $\hfill \Box$

\begin{table*}[!ht]
\begin{tabularx}{\textwidth}{@{}l*{15}{C}c@{}}
\toprule
% Group $\rightarrow$    & \multicolumn{4}{|@{}c@{\hskip0.25in}|}{Group 1} & \multicolumn{4}{@{}c@{\hskip0.25in}|}{Group 2}  & \multicolumn{3}{@{}c@{\hskip0.25in}}{Group 3} \\ 
% \midrule
\diagbox[width=3em]{\boldmath$\alpha$}{\boldmath$\beta$}  & \rotatebox[origin=c]{0}{\boldmath$0$}  & \rotatebox[origin=c]{0}{\boldmath$0.1$}  & \rotatebox[origin=c]{0}{\boldmath$0.2$}  & \rotatebox[origin=c]{0}{\boldmath$0.3$}  & \rotatebox[origin=c]{0}{\boldmath$0.4$}   & \rotatebox[origin=c]{0}{\boldmath$0.5$}  & \rotatebox[origin=c]{0}{\boldmath$0.6$}   & \rotatebox[origin=c]{0}{\boldmath$0.7$}  & \rotatebox[origin=c]{0}{\boldmath$0.8$}  & \rotatebox[origin=c]{0}{\boldmath$0.9$}  & \rotatebox[origin=c]{0}{1}  \\ \midrule
\boldmath$0.01$ & 122 \textbf{/} 145 & 121 \textbf{/} 144 & 124 \textbf{/} 141 & 126 \textbf{/} 134 & 129 \textbf{/} 136 & 127 \textbf{/} 136 & 132 \textbf{/} 138 & 127 \textbf{/} 133 & 130 \textbf{/} 131 & 132 \textbf{/} 138 & 137 \textbf{/} 135  \\
\boldmath$0.1$ & 121 \textbf{/} 149 & 122 \textbf{/} 148 & 121 \textbf{/} 149 & 125 \textbf{/} 146 & 121 \textbf{/} 145 & 136 \textbf{/} 132 & 136 \textbf{/} 137 & 135 \textbf{/} 136 & 135 \textbf{/} 133 & 130 \textbf{/} 131 & 134 \textbf{/} 131 \\ 
\boldmath$0.2$ & 124 \textbf{/} 144 & 121 \textbf{/} 143 & 124 \textbf{/} 140 & 124 \textbf{/} 147 & 134 \textbf{/} 139 & 137 \textbf{/} 139 & 144 \textbf{/} 134 & 129 \textbf{/} 138 & 138 \textbf{/} 138 & 138 \textbf{/} 130 & 134 \textbf{/} 133 \\ 
\boldmath$0.3$ & 138 \textbf{/} 137 & 127 \textbf{/} 142 & 133 \textbf{/} 133 & 135 \textbf{/} 136 & 138 \textbf{/} 138 & 138 \textbf{/} 140 & 131 \textbf{/} 144 & 140 \textbf{/} 138 & 123 \textbf{/} 144 & 127 \textbf{/} 150 & 126 \textbf{/} 141 \\
\boldmath$0.4$ & 133 \textbf{/} 138 & 142 \textbf{/} 134 & 136 \textbf{/} 135 & 129 \textbf{/} 137 & 136 \textbf{/} 139 & 138 \textbf{/} 140 & 130 \textbf{/} 141 & 139 \textbf{/} 136 & 128 \textbf{/} 147 & 127 \textbf{/} 148 & 126 \textbf{/} 147 \\ 
\boldmath$0.5$ & 141 \textbf{/} 135 & 138 \textbf{/} 135 & 144 \textbf{/} 133 & 141 \textbf{/} 138 & 142 \textbf{/} 137 & 136 \textbf{/} 139 & 138 \textbf{/} 139 & 127 \textbf{/} 148 & 138 \textbf{/} 140 & 125 \textbf{/} 147 & 129 \textbf{/} 143 \\ 
\boldmath$0.6$ & 138 \textbf{/} 131 & 141 \textbf{/} 134 & 141 \textbf{/} 134 & 142 \textbf{/} 132 & 142 \textbf{/} 137 & 134 \textbf{/} 142 & 129 \textbf{/} 142 & 134 \textbf{/} 141 & 130 \textbf{/} 144 & 124 \textbf{/} 146 & 130 \textbf{/} 146 \\
\boldmath$0.7$ & 143 \textbf{/} 133 & 138 \textbf{/} 141 & 138 \textbf{/} 138 & 139 \textbf{/} 138 & 139 \textbf{/} 138 & 137 \textbf{/} 138 & 134 \textbf{/} 141 & 131 \textbf{/} 143 & 136 \textbf{/} 140 & 129 \textbf{/} 146 & 129 \textbf{/} 147 \\ 
\boldmath$0.8$ & 143 \textbf{/} 135 & 138 \textbf{/} 137 & 141 \textbf{/} 137 & 144 \textbf{/} 133 & 142 \textbf{/} 137 & 142 \textbf{/} 137 & 141 \textbf{/} 138 & 143 \textbf{/} 135 & 129 \textbf{/} 146 & 122 \textbf{/} 148 & 126 \textbf{/} 147 \\ 
\boldmath$0.9$ & 144 \textbf{/} 133 & 139 \textbf{/} 139 & 142 \textbf{/} 135 & 144 \textbf{/} 133 & 142 \textbf{/} 137 & 140 \textbf{/} 139 & 143 \textbf{/} 136 & 128 \textbf{/} 148 & 131 \textbf{/} 146 & 121 \textbf{/} 149 & 122 \textbf{/} 150 \\
\boldmath$1$ & 143 \textbf{/} 132 & 144 \textbf{/} 132 & 144 \textbf{/} 132 & 145 \textbf{/} 134 & 145 \textbf{/} 134 & 137 \textbf{/} 137 & 134 \textbf{/} 141 & 135 \textbf{/} 139 & 121 \textbf{/} 149 & 121 \textbf{/} 149 & 128 \textbf{/} 147 \\
\bottomrule
\end{tabularx}
 \caption{\label{tab:reward_vs_cost} Variations of $f_{r}(\mathcal{S}, y)/f_{c}(\mathcal{S}, y)$ with $\alpha$ and $\beta$.}
\end{table*}

\section{Simulation Results}\label{sec: results}

The performance of \texttt{RAGA} is evaluated through extensive simulations with various environment maps and varying numbers and locations of nodes.\footnote{\texttt{RAGA} code is available at \url{https://github.com/rishabbala/Risk-Aware-TSP}}
%\lz{we need a better name for the alg}
% present  the results as obtained from empirical studies and evaluations of the Sequential Greedy Algorithm

\textbf{Simulation setup.} We consider the scenario with the number of sites (nodes) $|\mathcal{V}|=8$ in a 2D environment of size $100m\times 100m$ meters. %The $|\mathcal{V}|$ sites' positions are randomly selected in a square environment ${E}$ with side length $100$ meters. 
The sensing radius of the robot is $R=2$ meters. We assume both the reward and cost of an edge are modeled as a truncated Gaussian distribution, but \texttt{RAGA} can handle other distributions as well \textcolor{black}{since it only requires samples of the distributions to approximate CVaR. Similarly, the noise terms $y_{r}$ and $y_{c}$ are assumed to be Gaussian. However, the terms are generic to accommodate any other distributions.} Given the information density map $M$ (Fig.~\ref{fig:info_map}), in which each point depicts the mean reward obtained from observing that position. The average reward for an edge $r(e)$ is the sum of rewards of all points on ${M}$ sensed while traversing this edge. The mean cost $c(e)$ of an edge is proportional to its length. We assume the variances of the edge reward is proportional to its mean $r(e)$ and variance in the cost to be proportional to $C - c(e)$. The average rewards and costs for every edge are normalized and re-scaled to a maximum value of $10$. We set the number of samples as $n_{s}=250$. 
% We compute the reward $f_{r}(\mathcal{S}, y_{r})$ and the cost $f_{c}(\mathcal{S}, y_{c})$ of the tour $\mathcal{S}$ by summing the stochastic rewards and costs of all edges in it.The variance of $f_{r}(\mathcal{S}, y_{r})$ and $f_{c}(\mathcal{S}, y_{c})$ are then defined to be proportional to their respective means.

\textbf{Results.} Fig. \ref{fig:H for fixed beta} shows the probability distributions of the utility function $f(\mathcal{S}, y)$ as a function of $\alpha$, with a fixed $\beta$. %Consider, the plot 
From Fig. \ref{fig:beta=0.7},  when the risk level $\alpha$ is small, a path with a lower mean utility value $f(\mathcal{S}, y)$ is chosen. This is because a lower $\alpha$ indicates a small risk level, and therefore a low-utility low-risk path is selected. As the value of $\alpha$ increases, we see that the paths selected are more rewarding but tend to have a higher variance. \textcolor{black}{The case of $\alpha = 1$ is the risk-neutral scenario, where we disregard the variance in reward and cost but rather focus only on the mean values, which is the same as the deterministic settings in \cite{jawaid2013maximum}.} As expected, with $\alpha=1$, we select tours with the highest risk. This gradation in tour selection thus illustrates \texttt{RAGA}'s ability to select paths based on its evaluation of reward-risk trade-off.

Fig. \ref{fig:H for fixed alpha} shows the probability distributions of the utility function $f(\mathcal{S}, y)$ as a function of $\beta$, when $\alpha$ is given. Let us use Fig. \ref{fig:alpha=0.4} as an example. We see that initially, for $\beta=0$, \texttt{RAGA} selects a high-variance high-utility path. This is because, at $\beta=0$, we only consider maximizing the reward and disregard the cost of the tour. As we increase $\beta$, we see that \texttt{RAGA} slowly shifts towards paths with lesser utility (lesser risk), with the most conservative paths chosen around $\beta = 0.2\sim 0.4$. At this point, we wish to minimize cost while maximizing reward simultaneously, and therefore are more cautious of both terms. As $\beta$ continues to increase further, we see that the tours with higher utilities are chosen, which have a higher variance. Finally, at $\beta=1$, we consider only the cost minimization and select a tour with high-variance high-utility as the case of $\beta=0$. Table \ref{tab:reward_vs_cost} \textcolor{black}{provides a quantitative representation of Fig. \ref{fig:H for fixed beta} and \ref{fig:H for fixed alpha} by} showing the variations in $f_{r}(\mathcal{S}, y)$ and $f_{c}(\mathcal{S}, y)$ with respect to $\alpha$ and $\beta$.

% \lz{1. use one plot to explicitly explain this, say, for example, in Fig.-(b), when beta =0.4, ....2. Also, mention when alpha =1, that is Smith's results, then we can have a comparison with the existing research, which makes our results stronger. }
% \lz{what can we see from this plot, explicitly explain, otherwise, we cannot fully utilize the figure. }
\begin{figure}
     \centering
     \begin{subfigure}[b]{0.325\linewidth}
         \includegraphics[width=\linewidth]{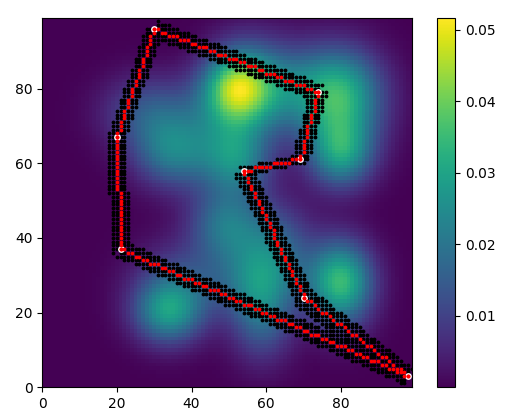}
         \caption{$\alpha=0.1, \beta=0$}
         \label{fig:path_0.1_0}
     \end{subfigure}
     \hfill
     \begin{subfigure}[b]{0.325\linewidth}
         \includegraphics[width=\linewidth]{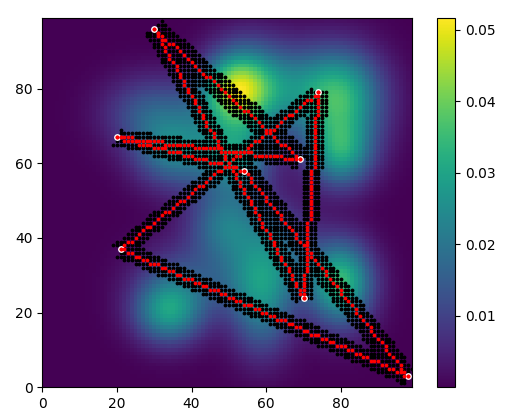}
         \caption{$\alpha=0.1, \beta=0.5$}
         \label{fig:path_0.1_0.5}
     \end{subfigure}
     \hfill
     \begin{subfigure}[b]{0.325\linewidth}
         \includegraphics[width=\linewidth]{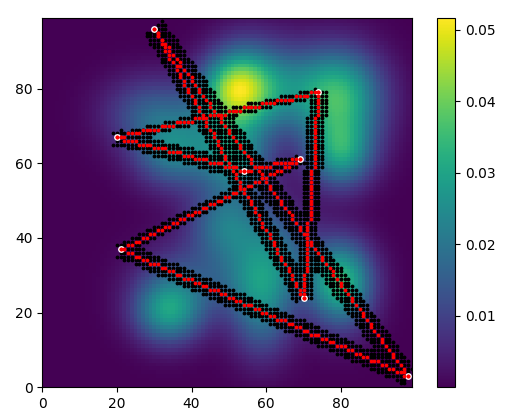}
         \caption{$\alpha=0.1, \beta=1$}
         \label{fig:path_0.1_1}
     \end{subfigure}
     \hfill\\
     \begin{subfigure}[b]{0.325\linewidth}
         \includegraphics[width=\linewidth]{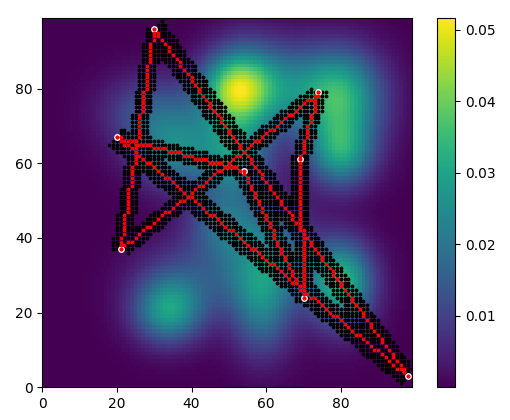}
         \caption{$\alpha=0.5, \beta=0$}
         \label{fig:path_0.5_0}
     \end{subfigure}
     \hfill
     \begin{subfigure}[b]{0.325\linewidth}
         \includegraphics[width=\linewidth]{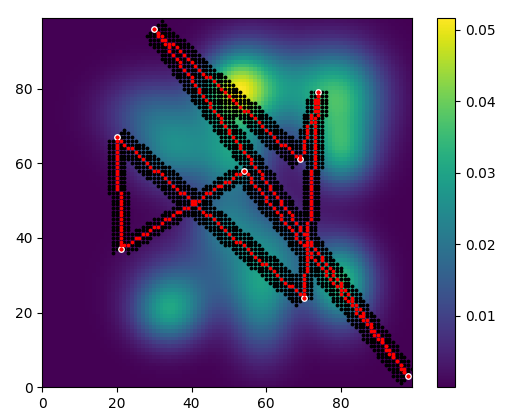}
         \caption{$\alpha=0.5, \beta=0.5$}
         \label{fig:path_0.5_0.5}
     \end{subfigure}
     \hfill
     \begin{subfigure}[b]{0.325\linewidth}
         \includegraphics[width=\linewidth]{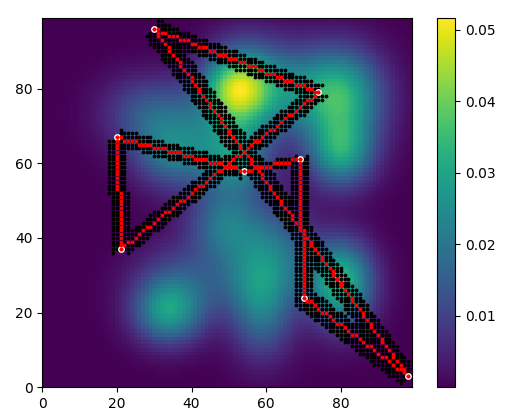}
         \caption{$\alpha=0.5, \beta=1$}
         \label{fig:path_0.5_1}
     \end{subfigure}
     \hfill\\
     \begin{subfigure}[b]{0.325\linewidth}
         \includegraphics[width=\linewidth]{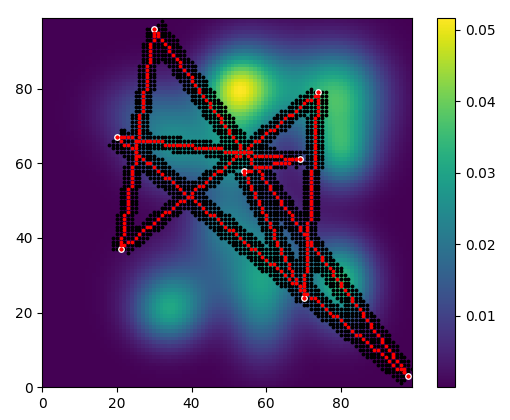}
         \caption{$\alpha=1, \beta=0$}
         \label{fig:path_1_0}
     \end{subfigure}
     \hfill
     \begin{subfigure}[b]{0.325\linewidth}
         \includegraphics[width=\linewidth]{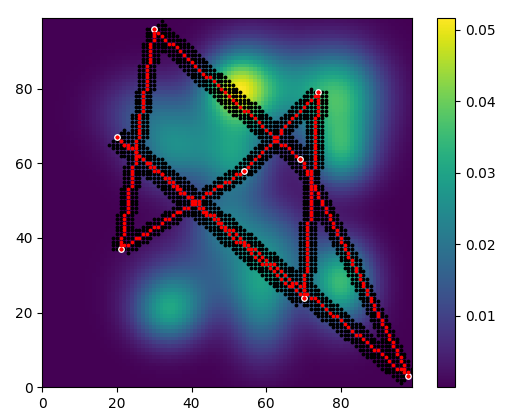}
         \caption{$\alpha=1, \beta=0.5$}
         \label{fig:path_1_0.5}
     \end{subfigure}
     \hfill
     \begin{subfigure}[b]{0.325\linewidth}
         \includegraphics[width=\linewidth]{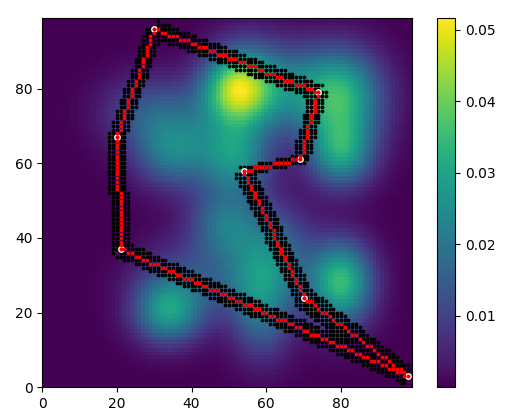}
         \caption{$\alpha=1, \beta=1$}
         \label{fig:path_1_1}
     \end{subfigure}\\
    \caption{The tours chosen for varying values of $\alpha$ and $\beta$.}
    \label{fig:tours}
    \vspace{-0.3cm}
\end{figure}
Fig. \ref{fig:tours} shows the tours chosen by \texttt{RAGA} for different values of $\alpha$ and $\beta$. We can clearly see that the paths chosen in Fig. \ref{fig:path_0.1_0} and Fig. \ref{fig:path_1_1} are the same and Fig. \ref{fig:path_0.1_1} and Fig. \ref{fig:path_1_0} are similar. This is because we get a lower reward when we travel less and a higher reward when we travel more, thus highlighting our problem's true dual nature. %\lz{this should be as the third graph. When reader reads the simulation setting up, it is natural to have the qualitative results first then the quantitative results.}

\begin{figure*}
\begin{multicols}{3}
\includegraphics[width=6cm]{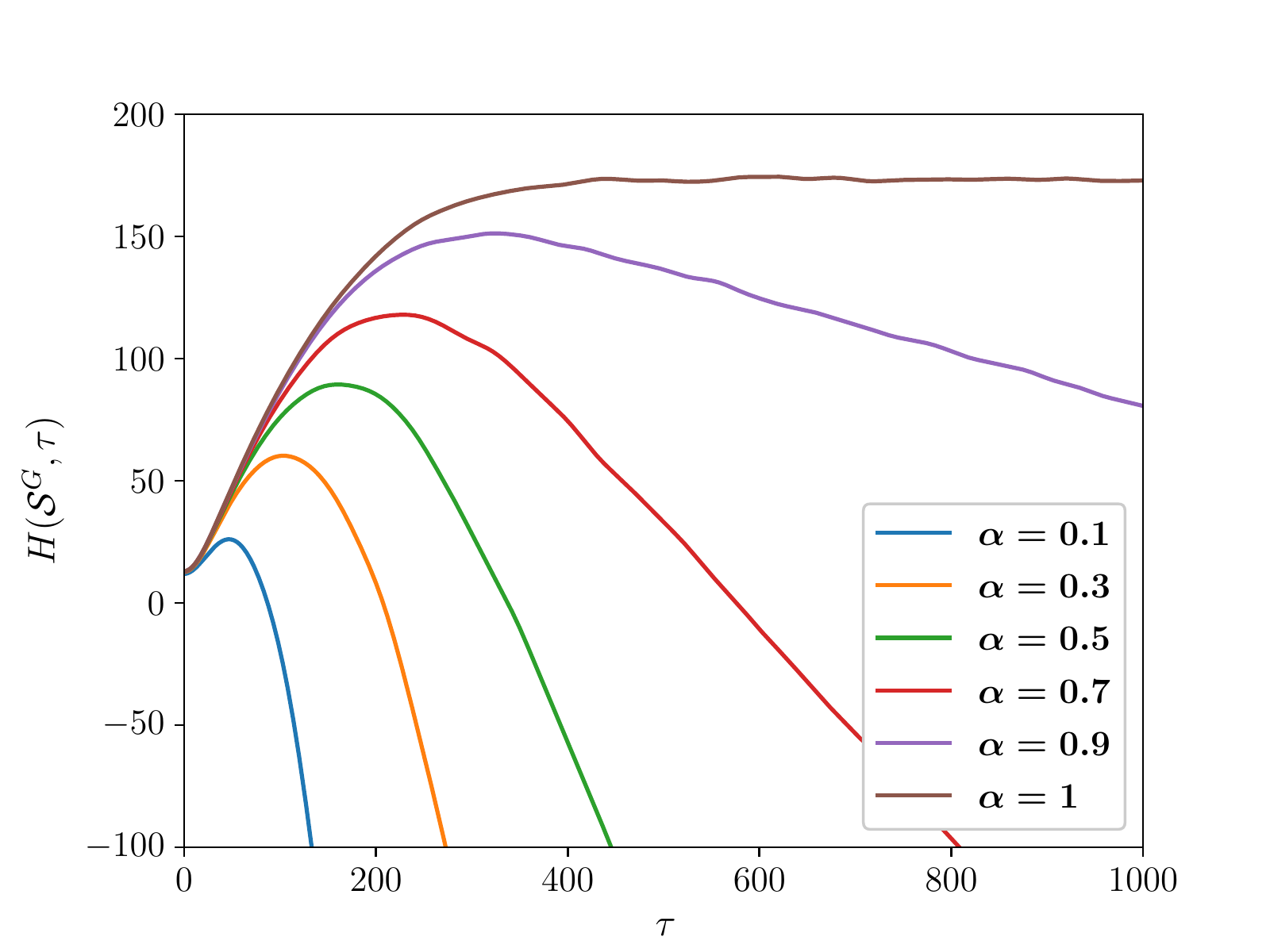}\subcaption{Variation of $H(\mathcal{S}, \tau)$ with respect to $\tau$}{}\label{fig:H_vs_tau}
    \includegraphics[width=6cm]{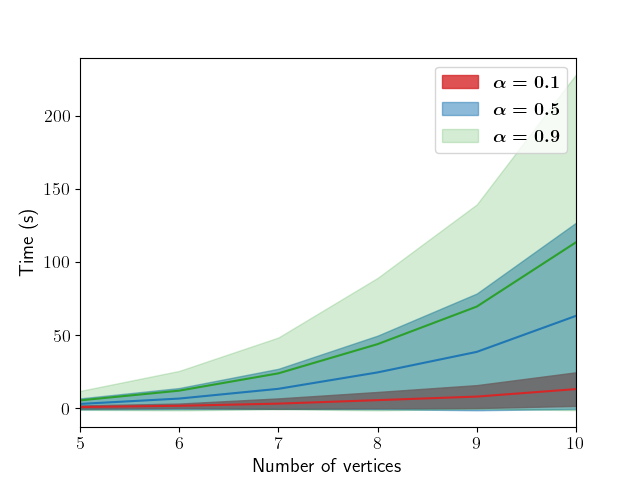}\subcaption{Runtime performance of \texttt{RAGA}}\label{fig:runtime}
    \includegraphics[clip, trim= 10mm 5mm 10mm 5mm, width=6cm]{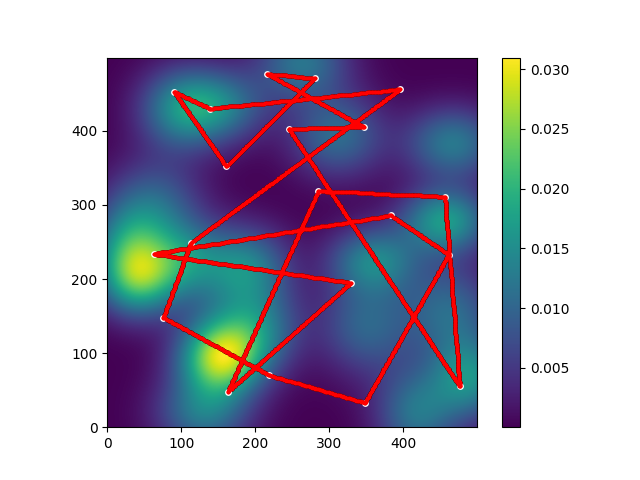}\subcaption{The \texttt{RAGA} run on a map with 20 nodes}\label{fig:20_nodes}
\end{multicols}
\vspace{-0.4cm}
\caption{Performance  of \texttt{RAGA}.}\label{fig:Performance}
\end{figure*}
% Fig. \ref{fig:3d} shows the distribution of the value of the utility function $f(\mathcal{S}, y)$, with respect to different values of $\alpha$ and $\beta$. It is clear that in this scenario, we get a higher value of $f(\mathcal{S}, y)$ when the tour cost is minimized, which is supported by Fig. \ref{fig:H for fixed alpha}, where our algorithm becomes most conservative around $\beta=0.4$. \lz{not clear, why?} Fig. \ref{fig:reward-var} shows the distribution of the reward-risk for three values of $\alpha$. It is as expected that for higher values of $\alpha$, a high-risk high-reward tour is selected. \lz{redundant, same as before?}

Fig. \ref{fig:H_vs_tau} shows the value of $H(\mathcal{S}, \tau)$ with respect to $\tau$, and it can be seen that the auxiliary function is concave. Fig. \ref{fig:runtime} shows the computational time for varying number of vertices for $\alpha = 0.1, 0.5, 0.9$. We use $n_{s}=250$, $\Gamma = 200$, and $\gamma = 1$. The results show that the running time increases with increase in $\alpha$ as for a larger $\alpha$, reaching the maximum is slower. This is shown in Fig. \ref{fig:H_vs_tau} where the maximum for $H(\mathcal{S}, \tau)$ is reached at a larger value of $\tau$ when $\alpha$ is larger. Combining this, and lines \ref{line:exit_start} - \ref{line:exit_end} in \texttt{RAGA}, we can see that with a smaller value of $\alpha$, \texttt{RAGA} stops quickly. As a result, the number of iteration over $\tau$ performed is lesser, and \texttt{RAGA} reaches a solution earlier than for a larger $\alpha$. To show the scalability of \texttt{RAGA}, we test it on an environment of size $500m\times 500m$ consisting of 20 nodes  along with the generated tour as shown in Fig. \ref{fig:20_nodes}. The parameters for the robot are set as $R =1$,  $\alpha=0.1$ and $\beta=0.8$.

\section{Conclusions And Future Work}\label{sec: conc}
In this paper, we developed a risk-aware greedy algorithm (\texttt{RAGA}) for the  TSP, while considering the risk-reward trade-off. We used a CVaR submodular maximization approach for selecting a solution set under matroidal constraints. We also analyzed the performance of \texttt{RAGA} and its running time. The results  show \texttt{RAGA}'s efficiency in optimizing both reward and cost simultaneously.

An ongoing work is to improve the running time of \texttt{RAGA} as a larger value of $\alpha$ has a higher running time. %We would like to overcome this, by replacing the ``\textit{for}" loop (Alg.~\ref{alg:riskg}, line \ref{line:for_start}) with an adaptive loop, where the value of $\tau$ changes based on the gradient of the auxiliary function, similar to gradient ascent. 
Another future avenue is to extend \texttt{RAGA} to address arbitrary distributions for the reward and cost, based on the real-world data. %Thirdly, we would like to extend our results to applications such as autonomous delivery and scheduling with multiple robots. 
Further, the work can be extended for other types of combinatorial optimization problems, and for the cases where the risk associated with the graph can be learned, and the paths can be determined dynamically. \textcolor{black}{Finally, we also plan to extend \texttt{RAGA} to the stochastic version of multi-robot TSP~\cite{frederickson1976approximation} where we plan tours for multiple robots with stochastic rewards and costs.}

\bibliographystyle{IEEEtran}
\bibliography{refs}

\end{document}